%% file: paper.tex
\documentclass[letterpaper, 10 pt, conference]{ieeeconf}  

\IEEEoverridecommandlockouts                              

\input{packages}
\input{commands}

\author{Antony Thomas and Giulio Ferro and
Fulvio Mastrogiovanni and Michela Robba
\thanks{$^\ast$Department of Informatics, Bioengineering, Robotics, and Systems Engineering, University of Genoa, Via All'Opera Pia 13, 16145 Genoa, Italy. \textit{antony.thomas@dibris.unige.it, giulio.ferro@unige.it, fulvio.mastrogiovanni@unige.it, michela.robba@unige.it}}}

\title{Computational Tradeoff in Minimum Obstacle Displacement Planning for Robot Navigation}

\begin{document}
\maketitle             
\begin{abstract}
In this paper, we look into the minimum obstacle displacement (MOD) planning problem from a mobile robot motion planning perspective. This problem finds an optimal path to goal by displacing movable obstacles when no path exists due to collision with obstacles. However this problem is computationally expensive and grows exponentially in the size of number of movable obstacles. This work looks into approximate solutions that are computationally less intensive and differ from the optimal solution by a factor of the optimal cost. 
\end{abstract}
\section{Introduction}
\input{introduction}
%
\section{Related Work}
\label{sec:related_work}
\input{related_work}

\section{Problem Definition}
\label{sec:problem_definition}
\input{problem_definition}
\section{The Minimum Obstacle Displacement Problem}
\label{sec:approach}
\input{approach}

%
\section{Horizon Slicing}
\label{sec:horizion_slicing}
\input{horizon_slicing}
\section{Evaluation}
\label{sec:results}
\input{results}

\section{Conclusion}
\label{sec:discussion}
\input{discussion}

\bibliographystyle{ieeetr}
\bibliography{references}
\end{document}

%% file: packages.tex
\usepackage[english]{babel}
\usepackage[utf8]{inputenc}

\usepackage{url}
\usepackage{makeidx}         
\usepackage{multicol}        
\usepackage[bottom]{footmisc}

\usepackage{amsmath,bm,amsfonts,amssymb}
\usepackage{amsthm}

\usepackage{commath}      

\usepackage{color,xcolor,ucs}
\usepackage{subfig}
\usepackage[font=small,labelfont=bf]{caption}

\usepackage{floatrow}
\usepackage{tabularx}
\usepackage{float}
\usepackage{cite}
\usepackage{xr-hyper}
\usepackage{wrapfig}

\usepackage{algorithm}
\usepackage{algpseudocode}
\usepackage{multirow}
\usepackage{listings}
\usepackage{lipsum}
\usepackage{textcomp} 

\usepackage[nocomma]{optidef}  

\usepackage{graphicx}
\usepackage{comment}
\usepackage[export]{adjustbox}

%% file: commands.tex
\theoremstyle{remark}
\newtheorem{definition}{Definition}
\newcommand{\B}{\mathbf}
\makeatletter
\newcommand{\clearsubcaptcounter}{\setcounter{sub\@captype}{0}}
\makeatother

%% file: introduction.tex
Classical robot motion planning approaches search for a feasible path from start to goal. Feasibility is often application specific and involve different constraints such as obstacle avoidance, mechanical limits, actuator limits, state uncertainties. A complete planner will search for all possible paths and if no feasible path exist due to constraint violation, the planner terminates reporting failure. However, in certain application domains it may be possible to alter some of the constraints to produce a feasible path. For example, manipulator robots often rearrange or move obstacles aside to complete a task, humanoid robots may need to move aside obstacles to reach a goal--- reposition chairs or open doors. When environmental and state uncertainties are represented by particles, finding the path that minimizes the probability of collision can be formulated as finding a path that collides with minimum number of particles~\cite{hauser2014IJRR}. 

Consider a humanoid robot in an office or home environment. While navigating through clutter we would envision the humanoid to exhibit human like navigation. A humanoid with such a capability will not always avoid obstacles but often rearrange or reposition movable obstacles such as coffee tables or chairs so as to synthesize a feasible path. Though there are different notions of optimality, we would want the humanoid to expend the minimum amount of \textit{work} while moving the obstacles. This can depend on a number of factors such as, the maximum forces available at the obstacle contact point, displacement of the contact point, number of obstacles moved. In this paper, we focus on the displacement factor, that is, \textit{how can a robot reach its goal by displacing obstacles while minimizing the total obstacle displacements} (overall displacement magnitudes). This also has applications in search and rescue wherein the robot minimizes debris displacement to reach the target quickly. \begin{figure}[t]
  \subfloat[]{\includegraphics[trim=50 121 40 100,clip,scale=0.6]{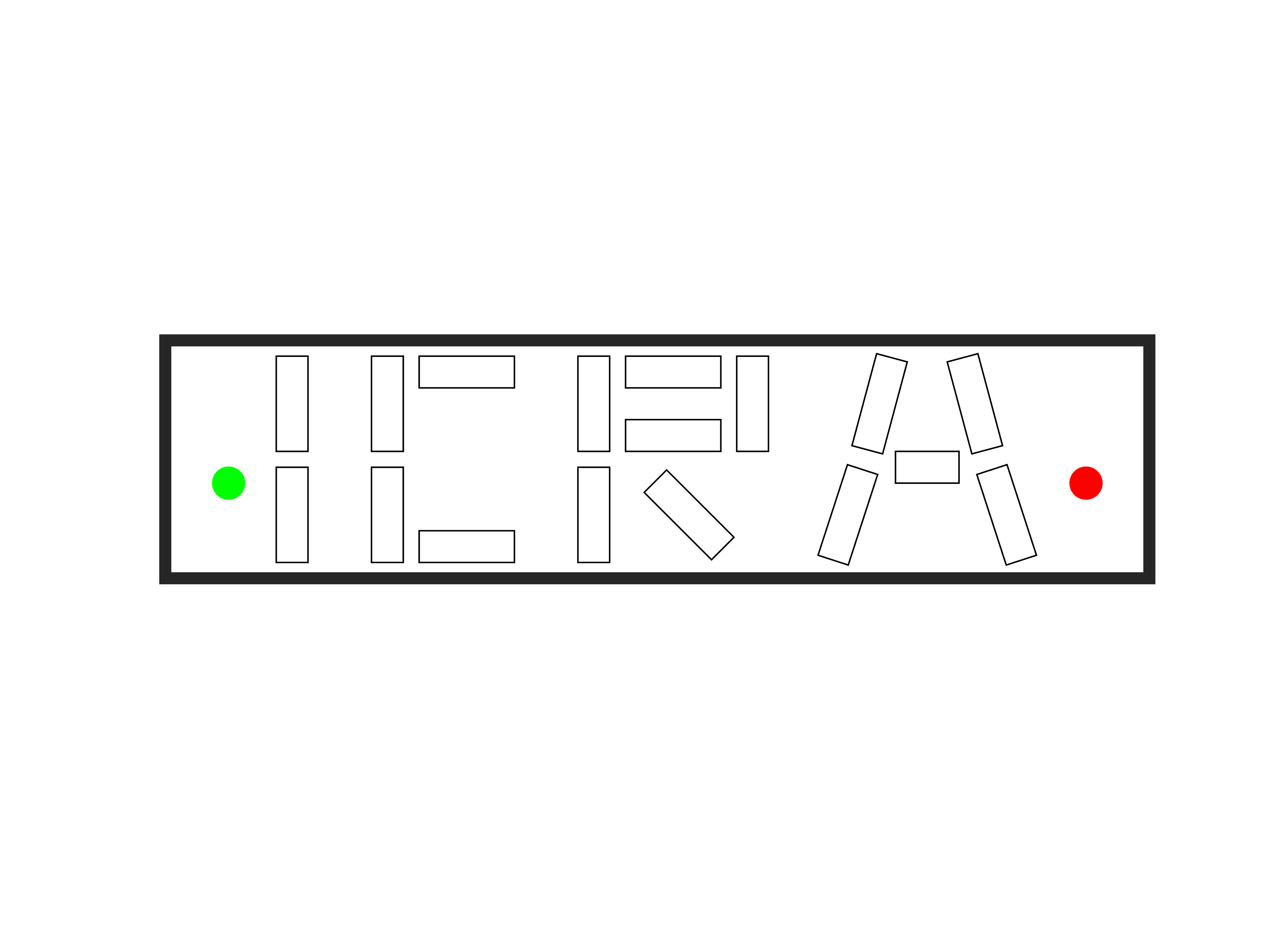}\label{fig:icra1}}\\
    \vspace{-0.4cm}
    \subfloat[]{\includegraphics[trim=50 121 40 100,clip,scale=0.59]{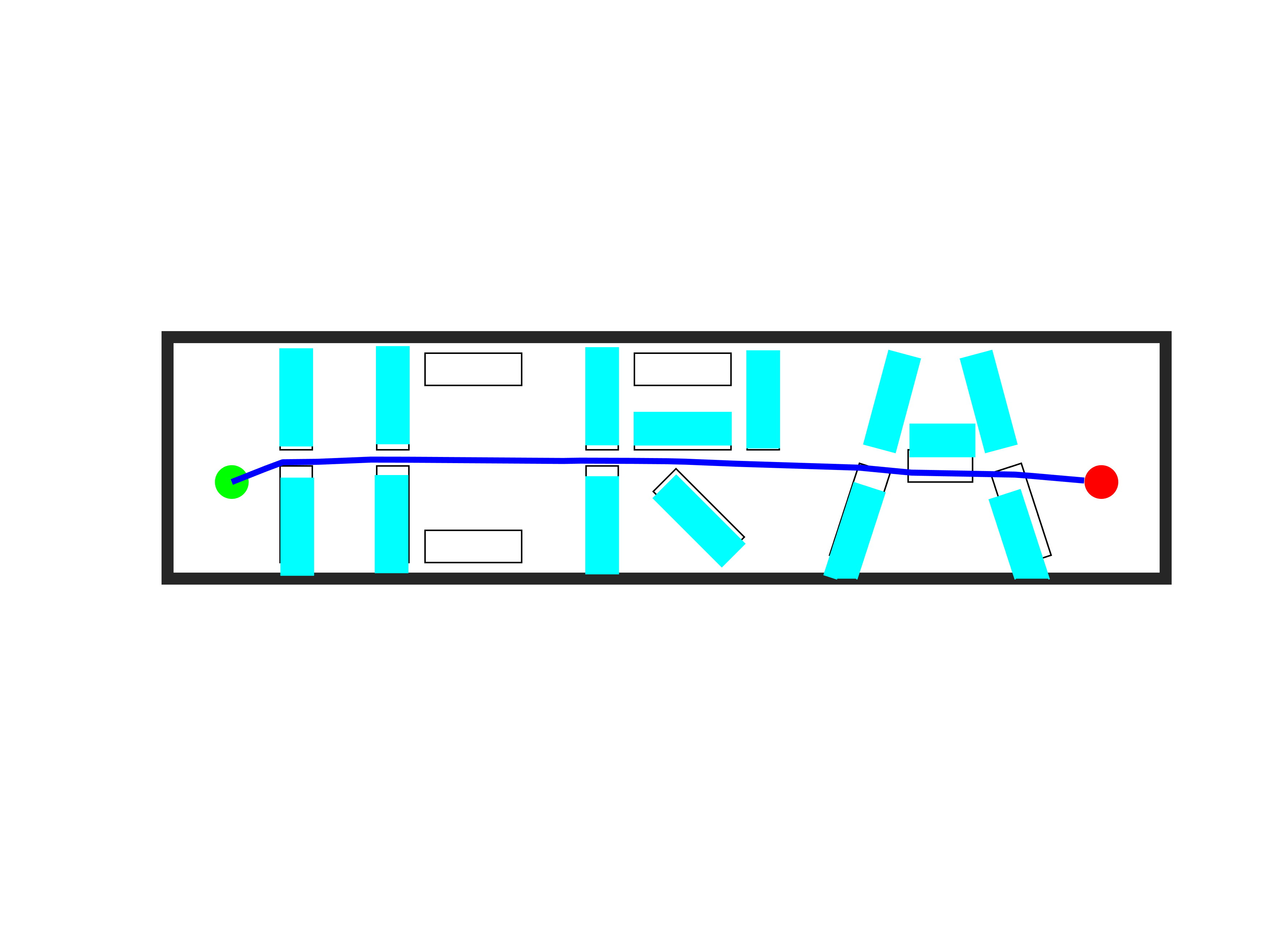}\label{fig:icra2}}\\
      \vspace{-0.4cm}
  \subfloat[]{\includegraphics[trim=50 121 40 100,clip,scale=0.6]{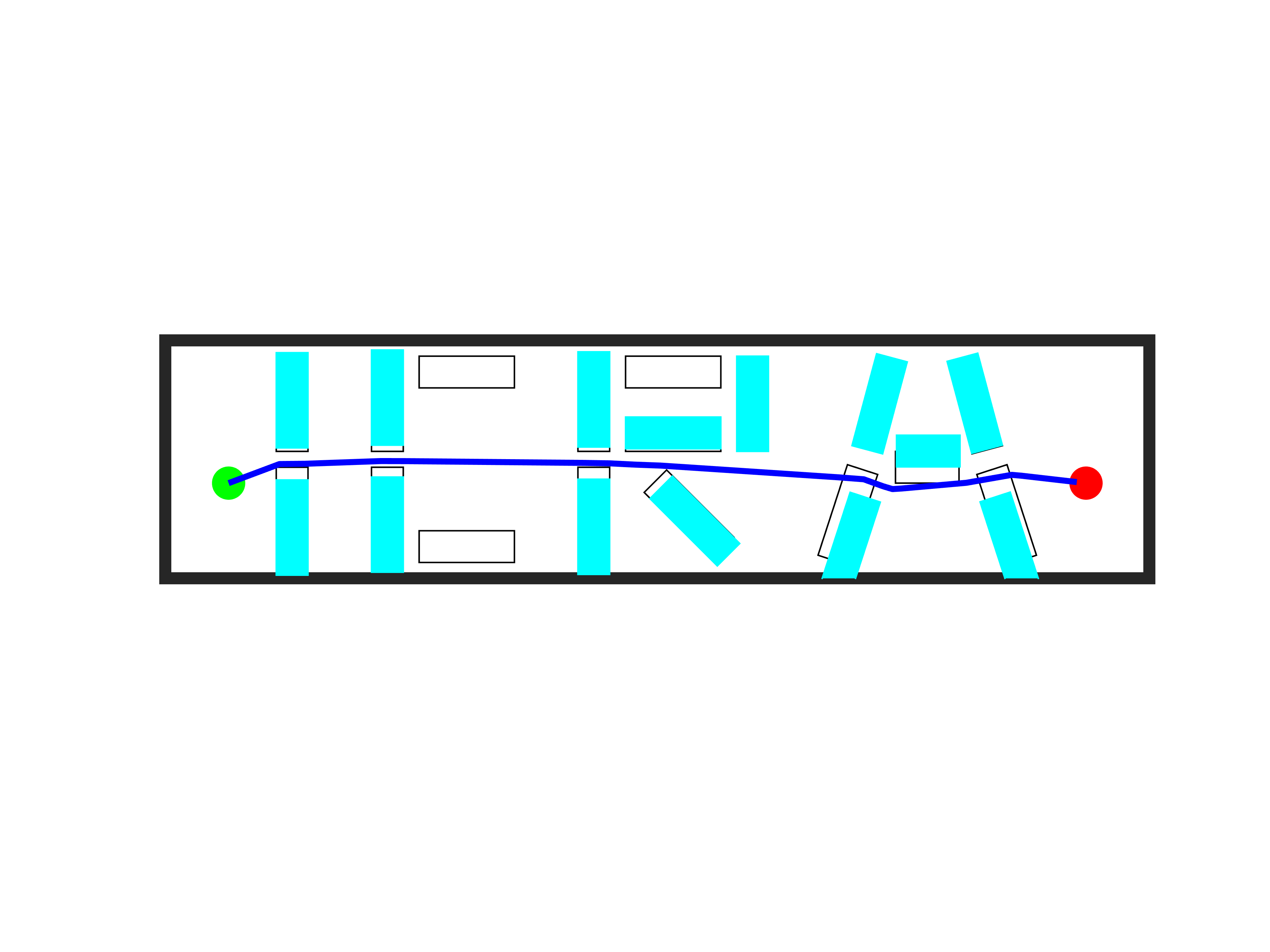}\label{fig:icra3}}
  \vspace{-0.2cm}
    \caption{(a) A robot asked to move from left to right in the ICRA domain with 17 movable obstacles (only translation allowed). (b) After 321 s, optimal solution is obtained with displaced obstacles shown in cyan. (c) An 0.24-optimal solution (details on $\epsilon$-optimality in Section~\ref{sec:horizion_slicing}) obtained in 25 s. }
  \label{fig:icradomain}
\end{figure}

\begin{figure}[t]
  \subfloat[]{\includegraphics[scale=0.23]{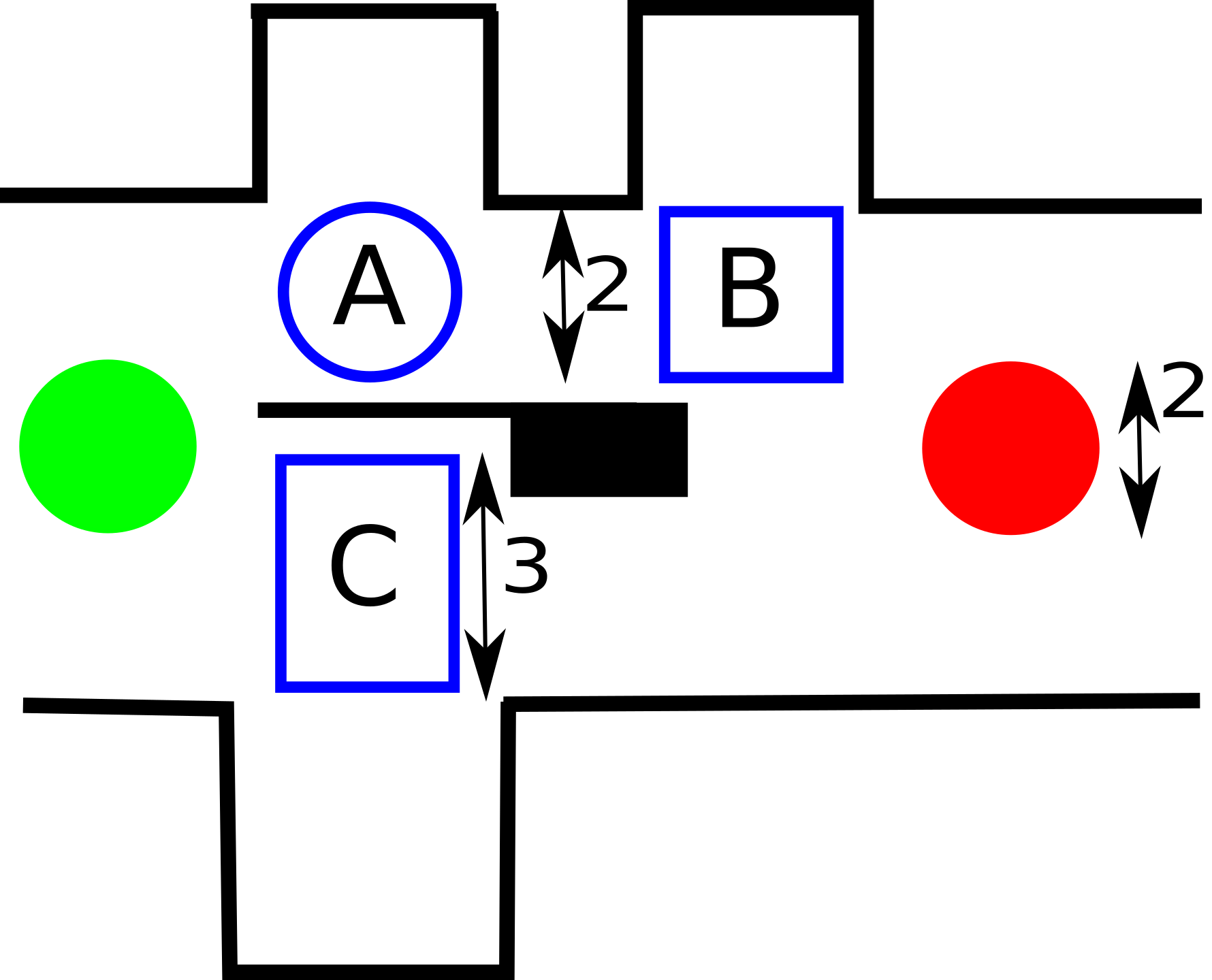}\label{fig:ex1}}\hfill
  \subfloat[]{\includegraphics[scale=0.23]{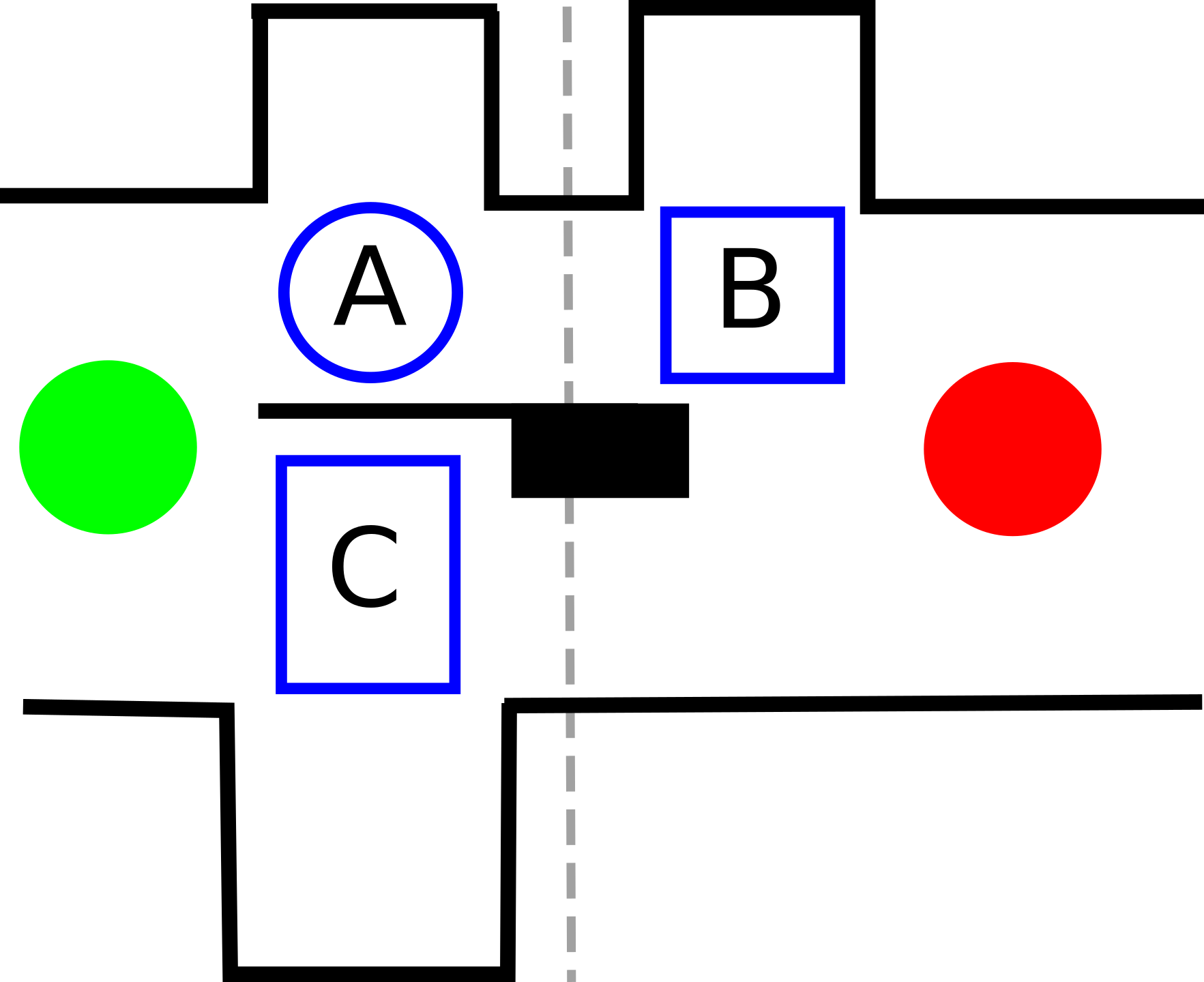}\label{fig:ex2}}\hfill
  \subfloat[]{\includegraphics[scale=0.23]{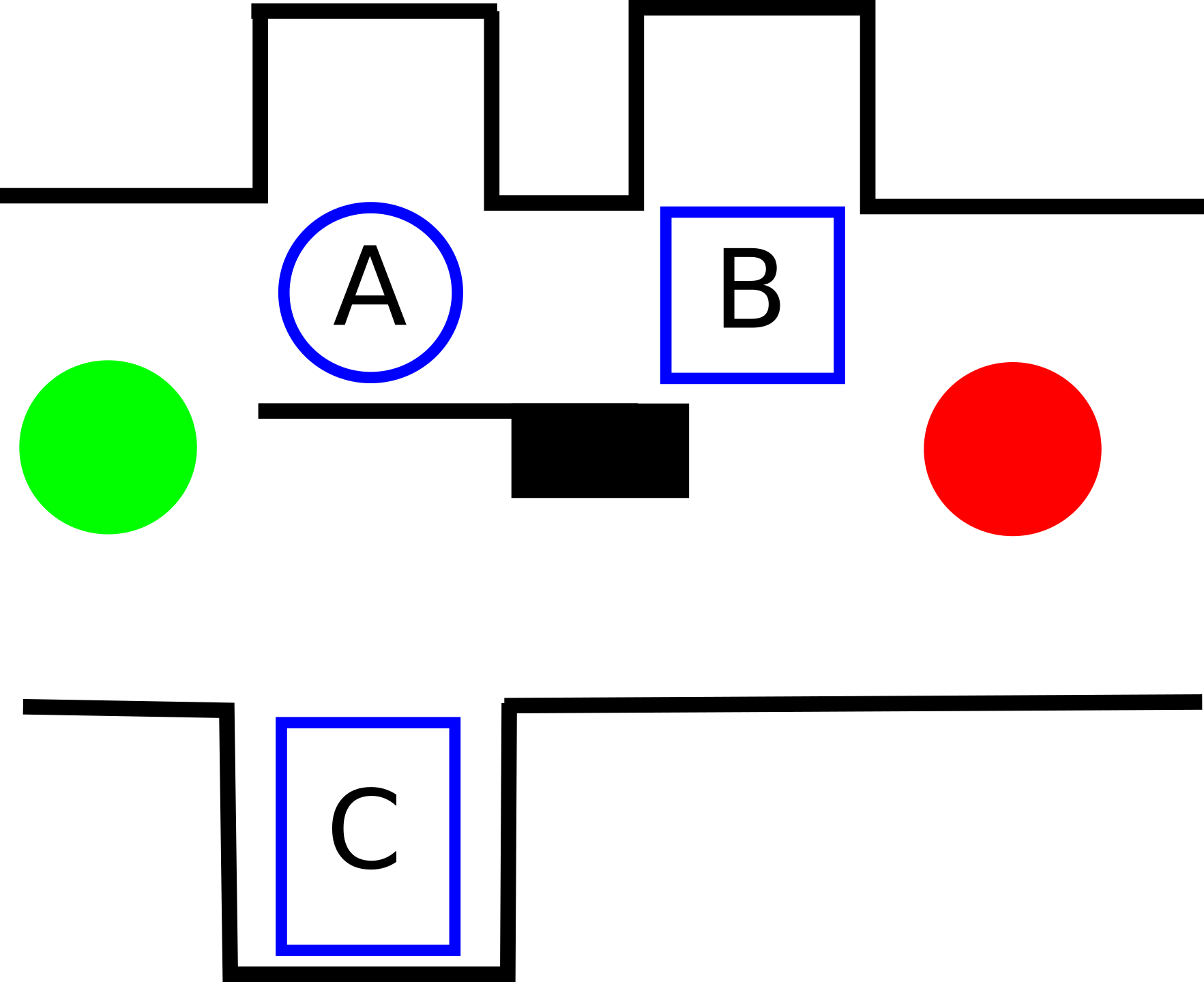}\label{fig:ex3}}
   \vspace{-0.2cm}
    \caption{(a) A simple domain where the robot is asked to move from left to right in the presence of three movable obstacles. (b) Solving the MOD problem considering two sub-problems (left and right of dashed line). This leads to a feasible path by moving obstacles A and B with total displacement magnitude of 4 units. (c) The optimal solution is obtained by displacing obstacle C by 3 units.}
  \label{fig:ex}
\end{figure}

In this paper, we formulate a Minimum Obstacle Displacement (MOD) problem that minimizes a weighted cost of robot path length and total obstacle displacement magnitudes as the robot navigates to its goal. This problem is NP-hard and therefore scenarios with even small number of obstacles can be computationally intensive (see Fig.~\ref{fig:icradomain}). Thus, it is appropriate to search for approximate methods and this work looks into one such method by dividing an MOD problem into sub-problems. This division into sub-problems is an approximation as MOD does not exhibit the optimal substructure property~\cite{cormen2009book}, that is, optimal solutions to sub-problems are not necessarily optimal. Consider a simple scenario shown in Fig.~\ref{fig:ex}. The robot is asked to move from left (green in figure) to right (red in figure) in the presence of movable obstacles (A, B and C in figure). The robot start and goal locations are chosen such that the robot path length for both the upper (displacing A and B) and the lower (displacing C) paths are equal in magnitude. In Fig.~\ref{fig:ex2}, the problem is split into two sub-problems. Since minimum displacement is solicited (path lengths equal as argued above), the robot chooses to move A by 2 units. For the next sub-problem, robot moves B by 2 units to reach the goal. Thus, the total displacement magnitude is 4 units. On the other hand, displacing C by 3 units clears a feasible path for the robot as shown in Fig.~\ref{fig:ex3} and is the optimal solution.  
The main contributions of the paper are: 1) Formulation and computation an of the optimal solution to the MOD problem, albeit computationally intensive, 2) Computation of approximate $\epsilon$-optimal solutions, that is, solutions that differ from the optimal solution by a factor of $\epsilon$. $\epsilon$-optimal solutions provide a very reasonable approximation with significant computational efficiency. Finally, in Section~\ref{sec:horizion_slicing}, we compute an upper bound for $\epsilon$. 

In this work, we are not concerned about how to move the obstacles, that is, we ignore the weight, size of the obstacles and assume that the computed displacements can be achieved by robot-obstacle or human-obstacle interaction. From an algorithmic perspective, this allows to perform the implementation on 2D projections of the 3D environment.



%% file: related_work.tex
This problem discussed herein is closely related to Navigation Among Movable Obstacles (NAMO) in~\cite{stilman2005IJHR,nieuwenhuisen2008WAFR,van2009WAFR}. NAMO class of problems are shown to be NP-hard if the
final locations of the obstacles are unspecified, and PSPACE-hard when the locations are specified~\cite{wilfong1991AMAI}. Most approaches thus solve a subclass of problems, focusing on efficiency. Manipulation among clutter or rearrangement planning in clutter~\cite{stilman2007ICRA,dogar2011RSS,krontiris2015RSS,karami2021AIIA} is another related class of problems wherein obstacles may need to be displaced to pick a target of interest. The Minimum Constraint Removal problem (MCR)~\cite{hauser2014IJRR} finds the minimum number of obstacles/constraints to be displaced to obtain a feasible path. Different algorithms for MCR exist in the literature~\cite{castro2013CDC,xu2016ASC,krontiris2017AR,xu2020ICMLC} and is proven to be NP-hard for convex polygonal obstacles~\cite{hauser2014IJRR}. Approaches for integrated task and motion planning~\cite{kaelbling2013IJRR,srivastava2014ICRA,dantam2016RSS,garrett2018IJRR,thomas2021RAS} also bears resemblance to the problem discussed in this paper since task execution may require removing (or adding) constraints at the motion planning level. Disconnection proving in motion planning~\cite{zhang2008WAFR,basch2001ICRA,li2021RSS} also pose related challenges since these approaches try to find constraints that prevents robot motion to goal.

The work in~\cite{thomas2022IAS} examine minimum displacement planning for movable obstacle through a 2 stage process. The first stage proceeds by finding a path through movable obstacle while minimizing robot-obstacle overlaps. The overlapping obstacles are then displaced iteratively by a distance factor until no overlap. Yet, the solutions returned are not necessarily optimal and no comment is made on the quality of the same. Probably the work that is more close to our approach is the Minimum Constraint Displacement (MCD) problem introduced by Hauser in~\cite{hauser2013RSS}. The approach in~\cite{hauser2013RSS} proceeds by (1) building a probabilistic roadmap of robot configurations and (2) sampling random obstacle displacements. The solution quality is improved by iteratively expanding the roadmap and the displacement samples. The method is shown to asymptotically approach the true optimum (which is unknown) with increasing robot configurations in the roadmap and with growing displacement samples. In contrast, we compute approximate solutions that deviates from the optimal solution cost by a factor of $\epsilon$. The approximate solutions offer significant computational advantage over the optimal solution. Further, MCD focus on pure motion planning in the sense that it ignores robot dynamics and other differential constraints and finds the translations and rotations required to move the robot. We plan a trajectory taking into account the robot dynamics and other constraints such as control limits, obstacle avoidance, to find the sequence of control actions that moves the robot to the goal.

%% file: problem_definition.tex
Throughout this paper we shall denote vectors by bold lower case letters, that is $\B{x}$ and its components by lower case letters. The transpose of $\B{x}$ will be denoted by $\B{x}^T$ and its Euclidean norm by $\norm{\B{x}} = \sqrt{\B{x}^T\B{x}}$. Sets will be denoted using mathcal fonts, that is, $\mathcal{S}$ or enclosed within braces ($\{\cdot\}$) and its cardinality will be denoted by $|\mathcal{S}|$. 


Let $\B{x}_k$ denote the robot state at any time $k$. For instance, in mobile robot navigation $\B{x}_k$ may denote the robot pose at time $k$. We consider a standard motion model for the robot given by $\B{x}_{k+1} = f(\B{x}_k,\B{u}_k)$,
\noindent where $\B{u}_k$ is the applied control action at time $k$. Further, let $\B{x}^s$ denote the start state and $\B{x}^g$ denote the goal state. Let $\mathcal{O}= \{\B{o}^i| \ 1 \leq i \leq |\mathcal{O}|\}$ denote the set of movable obstacle in the environment. By abuse of notation we will use $\B{o}^i$ to denote both the $i$-th obstacle as well as its state. The obstacles are associated with a displacement set $\mathcal{D}= \{\B{d}^i \in \mathbb{R}^n| \ 1 \leq i \leq |\mathcal{O}|\}$ that represents the corresponding obstacle displacements. The new obstacle location after being displaced by $\B{d}^i$ will be denoted by $\B{o}^i(\B{d}^i)$. Since the obstacles can either rotate or translate or perform both, $\B{d}^i$ belongs to a displacement space of arbitrary dimension. We can now define the MOD planning problem.
\begin{definition}
The Minimum Obstacle Displacement (MOD) planning problem finds a sequence of control actions $\B{u}_1,\ldots, \B{u}_{T-1}$ that navigates the robot from $\B{x}^s$ to $\B{x}^g$ with obstacle displacements $\B{d}^1,\ldots,\B{d}^n$ while minimizing the cost $C(\B{x},\mathcal{D})$ given by
\vspace{-0.3cm}
\begin{equation}
C = w^xc(\B{x}) + w^d \sum_{i=1}^{|\mathcal{O}|} c^i(\B{d}^i)
\label{eq:mdmp}
\end{equation}
\noindent such that 1) $\B{x}_1 = \B{x}^s$, 2) $\B{x}_T = \B{x}^g$, and 3) $\B{x}_k \cap \bigcup_{i=1}^n \B{o}^i(\B{d}^i) = \{\emptyset\}$, $\forall k \in \{1,\ldots, T\}$. 
\label{def:mdmp}
\end{definition}
In~(\ref{eq:mdmp}), $c(\B{x})$ is a function of robot location (say, path length), $c^i(\B{d}^i)$ is a function of displacement magnitudes and $w^x$, $w^d$ are the respective weights. Constraints 1 and 2 satisfies the endpoint constraints. The last condition guarantees that the robot does not intersect with the displaced obstacles. 

A straightforward but exhaustive approach is to consider all the $2^{|\mathcal{O}|}$ subsets of the movable obstacles with different displacements for each subset and finally selecting the minimum from among them. Note that the subsets for which there exists a solution (not necessarily optimal) are also the different set of obstacles such that removing them from the workspace connects the start and the goal. In fact, by setting $w^x=0$ and restricting $d^i$ to two values, $d^i=0$ for obstacle being present and $d^i=1$ for obstacle removed, MOD solves the MCR problem which is proved to be NP-hard~\cite{hauser2014IJRR}. Therefore, by reduction, MOD is NP-hard. 

%% file: approach.tex
In this section we present an exact algorithm for MOD which gives optimal solutions, albeit computationally expensive as the set $\mathcal{O}$ and the robot workspace grows. 
\subsection{Modeling Obstacle Displacement}
As noted before, in this work we are not concerned about how to move the obstacles and assume that obstacles can be displaced irrespective of the nature of the robot-obstacle interaction required. Thus, the obstacle may be moved without any constraints. Each obstacle is capable of translation and rotation and hence the state $\B{o}^i = [o^{i,x},o^{i,y},o^{i,\theta}]$ includes translations in the $x$ and $y$ directions and a rotation (the superscript $i$ will be often dropped to avoid clutter). This gives us the following linear model
\begin{equation}
    \begin{split}
        o^{x}_{k+1} &= o^{x}_k + \tau s^x_k\\
         o^{y}_{k+1} &= o^{y}_k + \tau s^y_k\\
          o^{\theta}_{k+1} &= o^{\theta}_k + \tau s^{\theta}_k\\
    \end{split}
    \label{eq:obs_dyn}
\end{equation}
\noindent where $\tau$ is the duration of a time-step and $\B{s}^i_k = [s^{i,x}_k, s^{i,y}_k, s^{i,\theta}_k]$ specify the velocities. Thus, $[\tau s^x_k, \tau s^y_k, \tau s^{\theta}_k]$ is the displacement between two time-steps for the i-th obstacle and the overall displacement $\B{d}^i =  [\sum_{k=1}^{T-1} \tau s^x, \sum_{k=1}^{T-1}\tau s^y, \sum_{k=1}^{T-1}\tau s^{\theta}]$. We may write~\eqref{eq:obs_dyn} compactly as $\B{o}_{k+1} = g(\B{o}_{k},\B{s}_k)$. 
\subsection{Robot Model} 
In general we may consider any standard motion model for the robot. In this paper, we consider two different linear models. The first is a trivial model similar to the obstacle model in~\eqref{eq:obs_dyn}
\vspace{-0.1cm}
\begin{equation}
    \begin{split}
   x_{k+1} &= x_k + \tau u^x_k\\
         y_{k+1} &= y_k + \tau u^y_k\\
          \theta_{k+1} &= \theta_k + \tau u^{\theta}_k\\
    \end{split}
    \label{eq:robot_model1}
\end{equation}
\noindent where $\B{x}_k = [x_k,y_k,\theta_k]$ is the robot pose at time $k$ and $\B{u}_k = [u^x_k,u^y_k,u^{\theta}_k]$ is the applied control. We also consider the linear dynamics given in~\cite{van2011lqg} with state $\B{x}_k = [x_k,y_k,v^x_k,v^y_k]$ consisting of its location and velocity with acceleration input $\B{u}_k= [a^x_k,a^y_k]$. The corresponding model is
\vspace{-0.1cm}
\begin{equation}
    \begin{split}
        x_{k+1} &= x_k + \tau v^x_k + (\tau^2/2)a^x_k \\
        y_{k+1} &= y_k + \tau v^y_k + (\tau^2/2)a^y_k  \\
        v^x_{k+1} &= v^x_k + \tau a^x_k \\
        v^y_{k+1} &= v^y_k + \tau a^y_k \\
    \end{split}
    \label{eq:robot_model2}
\end{equation}
\noindent where $\tau$ is the duration of a time step. The use of linear models allow to obtain global optimal solutions, albeit dependent on the nature of the objective function and other constraints. 

\subsection{The optimization problem}
We formulate the MOD problem as an optimization problem that finds a feasible path for the robot, minimizing the path length and obstacle displacement magnitudes. While doing so, the control inputs and state variables must lie within their respective feasible sets and the robot should not collide with the obstacles. The overall optimization problem is thus formalized as
\vspace{-0.1cm}
\begin{mini!}|s|[2]
{ }{\alpha^r J^r+ \alpha^o J^o}{\label{eq:optimization_problem}}{\label{eq:cost_fn0}}
  \addConstraint{\B{x}_{k+1}=f( \B{x}_k,\B{u}_k ) }{\label{eq:cost_fn1}} 
  \addConstraint{\B{o}_{k+1}=g( \B{o}_k,\B{s}_k ) }{\label{eq:cost_fn2}}
  \addConstraint{\underline{\B{x}}  \le \B{x}_{k} \le \overline{\B{x}} }{\label{eq:cost_fn3}}
  \addConstraint{\underline{\B{u}}  \le \B{u}_{k} \le \overline{\B{u}} }{\label{eq:cost_fn4}}
   \addConstraint{\underline{\B{o}}  \le \B{o}_{k} \le \overline{\B{o}} }{\label{eq:cost_fn5}}
   \addConstraint{\underline{\B{s}}  \le \B{s}_{k} \le \overline{\B{s}} }{}{\label{eq:cost_fn6}}
   \addConstraint{\norm{\B{x}_k-\B{o}_k^i }\geq r^r + r^o}{}{\label{eq:cost_fn7}}
      \addConstraint{\forall k \in \{1,\ldots,T-1\} \ \text{and} \ \forall i \in \{1,\ldots,|\mathcal{O}|\}} \nonumber
       \end{mini!}
       \vspace{-0.1cm}
\noindent where
\vspace{-0.1cm}
\begin{equation}
    J^{r}= \sum_{k = 1}^{T-1} \  \norm{\B{x}_{k+1}-\B{x}_{k}}^2  \ + \ \alpha^{re}\norm{\B{x}_T - \B{x}^g}^2 \label{obj_x}
    \end{equation}
    \noindent the first term minimizes the robot path length and the second term drives the robot to the goal with $\alpha^{re}$ being its weight and
\begin{equation}
    J^{o}= \sum_{k=1}^{T-1}\sum_{i = 1}^{|\mathcal{O}|} \ \  \tau \norm{\B{s}^i_k}^2 \label{obj_u_o}
\end{equation}
\noindent measures the obstacle displacement magnitudes. The weights $\alpha^r, \alpha^o$ result in relative trade-off between longer robot paths and larger obstacle displacements. The robot and obstacle dynamics are followed in constraints~\eqref{eq:cost_fn1}-\eqref{eq:cost_fn2} subject to the upper and lower bounds on the state and control variables as given in~\eqref{eq:cost_fn3}-\eqref{eq:cost_fn6}. Finally~\eqref{eq:cost_fn7} represents the collision avoidance constraint between the robot and the obstacles with $r^r$, $r^o$ denoting the robot and obstacle radii, respectively. The endpoint conditions, that is, $\B{x}_1 = \B{x}^s$ and $\B{x}_T = \B{x}^g$ are omitted to avoid further clutter. We note here that in Definition~\ref{def:mdmp}, $c(\B{x})$ and $c^i(\B{d}^I)$ where defined to be general functions of robot location and obstacle displacement, respectively. Comparing the above formulation with~\eqref{eq:mdmp} in Definition~\ref{def:mdmp}, it is easily identified that, $ \alpha^r = w^x$, $ \alpha^o = w^d$, $J^r = c(\B{x})$ and $J^o = \sum_{i=1}^{|\mathcal{O}|}c^i(\B{d}^i)$. For brevity, the objective function~\eqref{eq:cost_fn0} will be denoted using $J = \alpha^r J^r+ \alpha^o J^o$. 
\subsection{Collision Constraint}
At this point, it is essential to spend a few words regarding the modeling of the collision avoidance constraint in~\eqref{eq:cost_fn7}.
When the robot model~\eqref{eq:cost_fn1} is linear the optimization problem~\eqref{eq:optimization_problem} is a non-convex nonlinear program (NLP). However, the only element of non-convexity is the collision avoidance constraint in~\eqref{eq:cost_fn7} (all the other constraints are linear).
We thus linearize the collision avoidance constraint by introducing $j$ binary auxiliary variables $\xi^{j,i}_k \in \left \{0,1\right \}$ for each obstacle $\B{o}^i \in \mathcal{O}$ at every time step $k$. The constraint \eqref{eq:cost_fn7} is thus rewritten as
\begin{small}
\begin{align}
    x_k - o^{i,x}_k \geq r^r + r^o - \xi^{1,i}_k M \label{eq:linear_constraints1}\\
        - \left( x_k - o^{i,x}_k \right) \geq r^r + r^o - \xi^{2,i}_k M \label{eq:linear_constraints2} \\
         y_k - o^{i,y}_k \geq r^r + r^o - \xi^{3,i}_k M \label{eq:linear_constraints3}\\
          -\left( y_k - o^{i,y}_k \right) \geq r^r + r^o - \xi^{4,i}_k M \label{eq:linear_constraints4} \\
          \sum_{j=1}^4 \xi^{j,i}_k \leq 3 \label{eq:linear_constraints5}
          \end{align}
\end{small}
\noindent $\forall k \in \{1,\ldots,T-1\} \ \text{and} \ \forall i \in \{1,\ldots,|\mathcal{O}|\}$, where $M$ is an appropriate large constant. Thus by rewriting the single constraint~\eqref{eq:cost_fn7} into linear constraints~\eqref{eq:linear_constraints1}-\eqref{eq:linear_constraints5}, the optimization problem in~\eqref{eq:optimization_problem} is converted into an Mixed Integer Quadratic Program (MIQP). MIQP is a class of optimization problems with a quadratic objective function subject to linear constraints and involving both continuous and discrete decision variables. It is well known from operations research theory that there are algorithms that guarantee global optimality for this type of optimization problems. Nonlinear and non-convex robot models provide weak theoretical guarantees and the optimization often results in local minima. Thus, in such settings, a linear model may be employed first to find the required path. A trajectory tracking algorithm can then be used to compute the controls for the nonlinear model. 


%% file: horizon_slicing.tex
\begin{figure*}[]
  \subfloat[]{\includegraphics[trim=5 30 10 20,clip,scale=0.28]{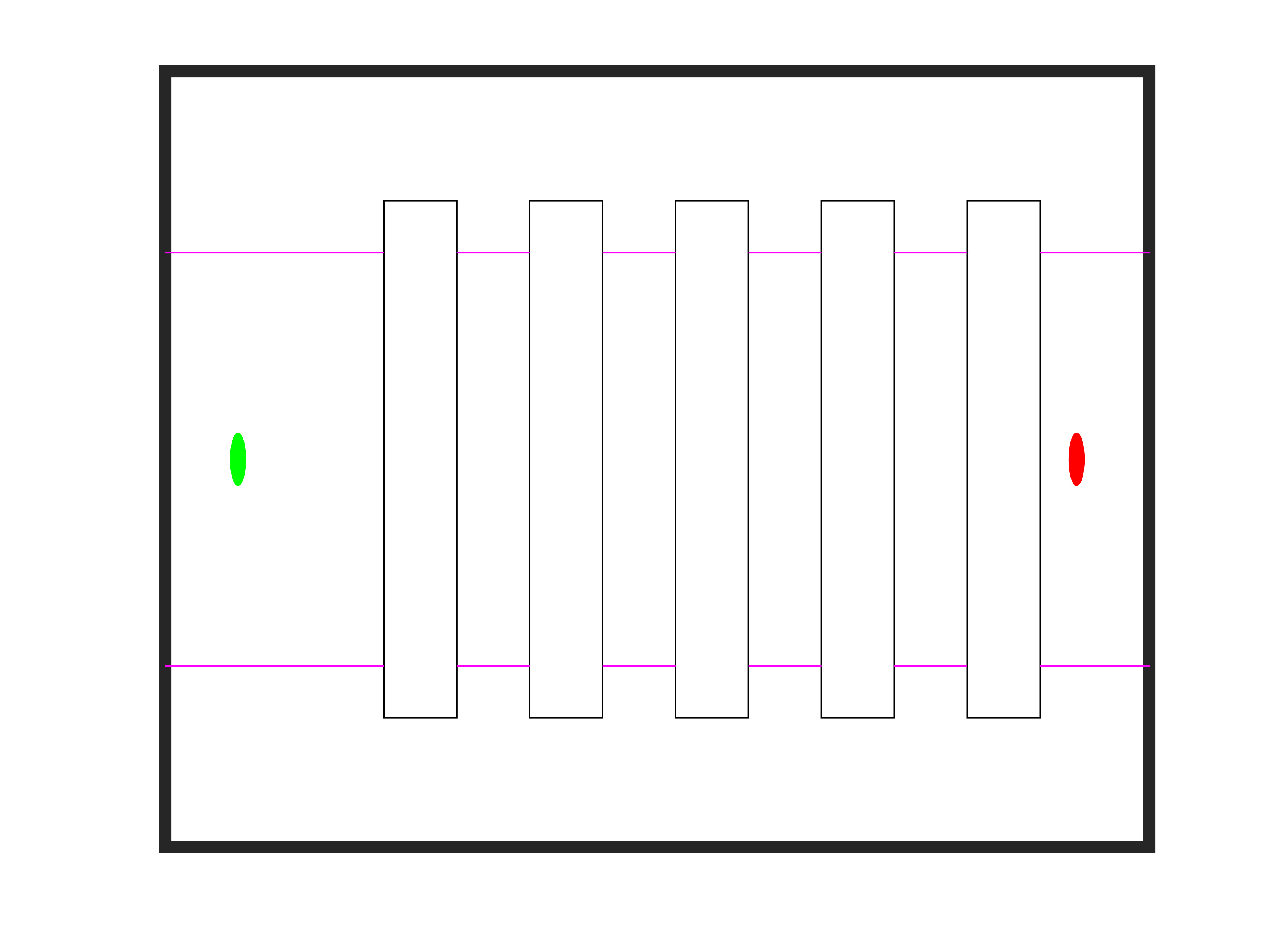}\label{fig:rec1}}
    \subfloat[]{\includegraphics[trim=5 30 10 20,clip,scale=0.28]{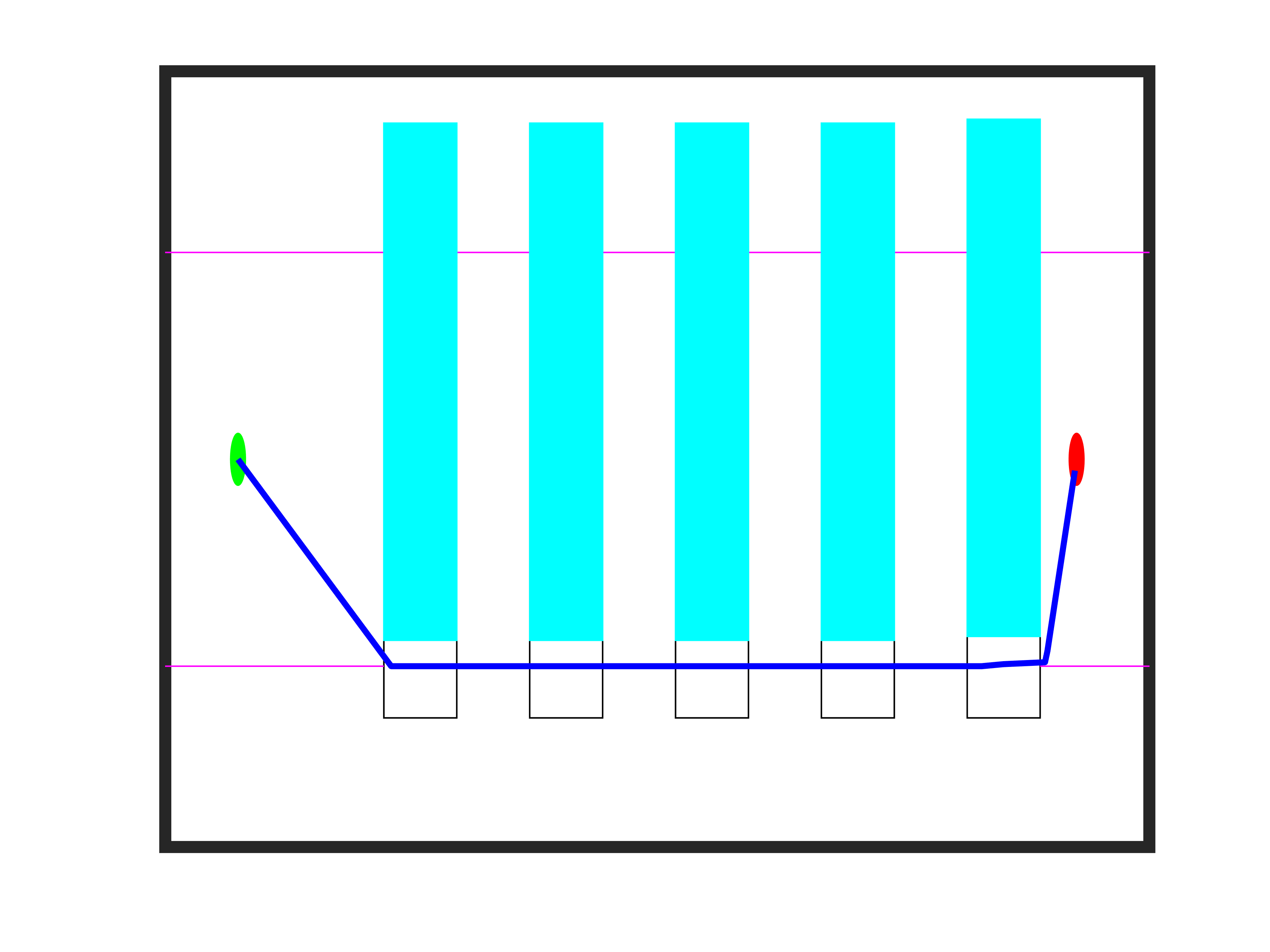}\label{fig:rec2}}
  \subfloat[]{\includegraphics[trim=5 30 10 20,clip,scale=0.28]{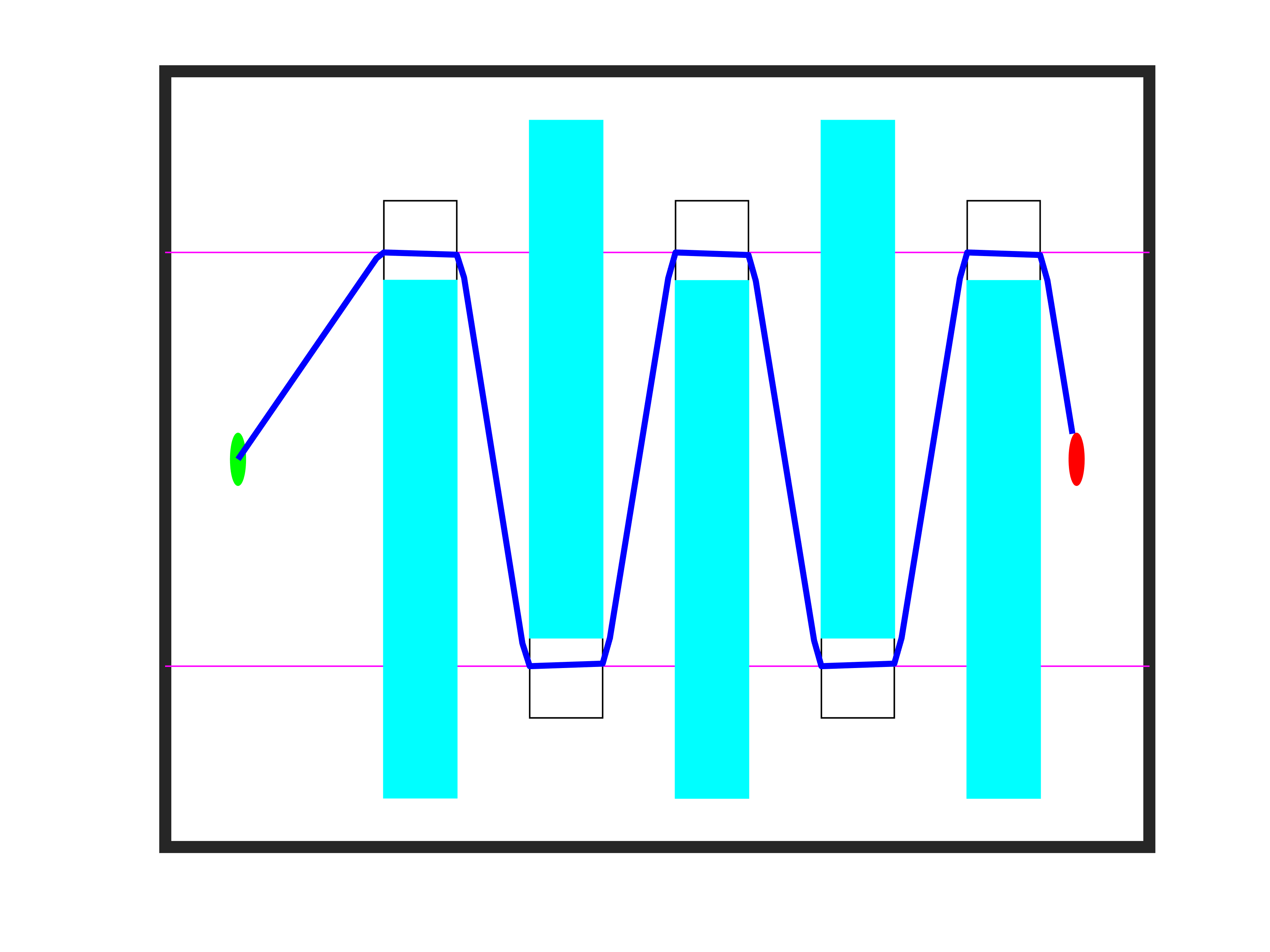}\label{fig:rec3}}
    \subfloat[]{\includegraphics[trim=5 30 10 20,clip,scale=0.28]{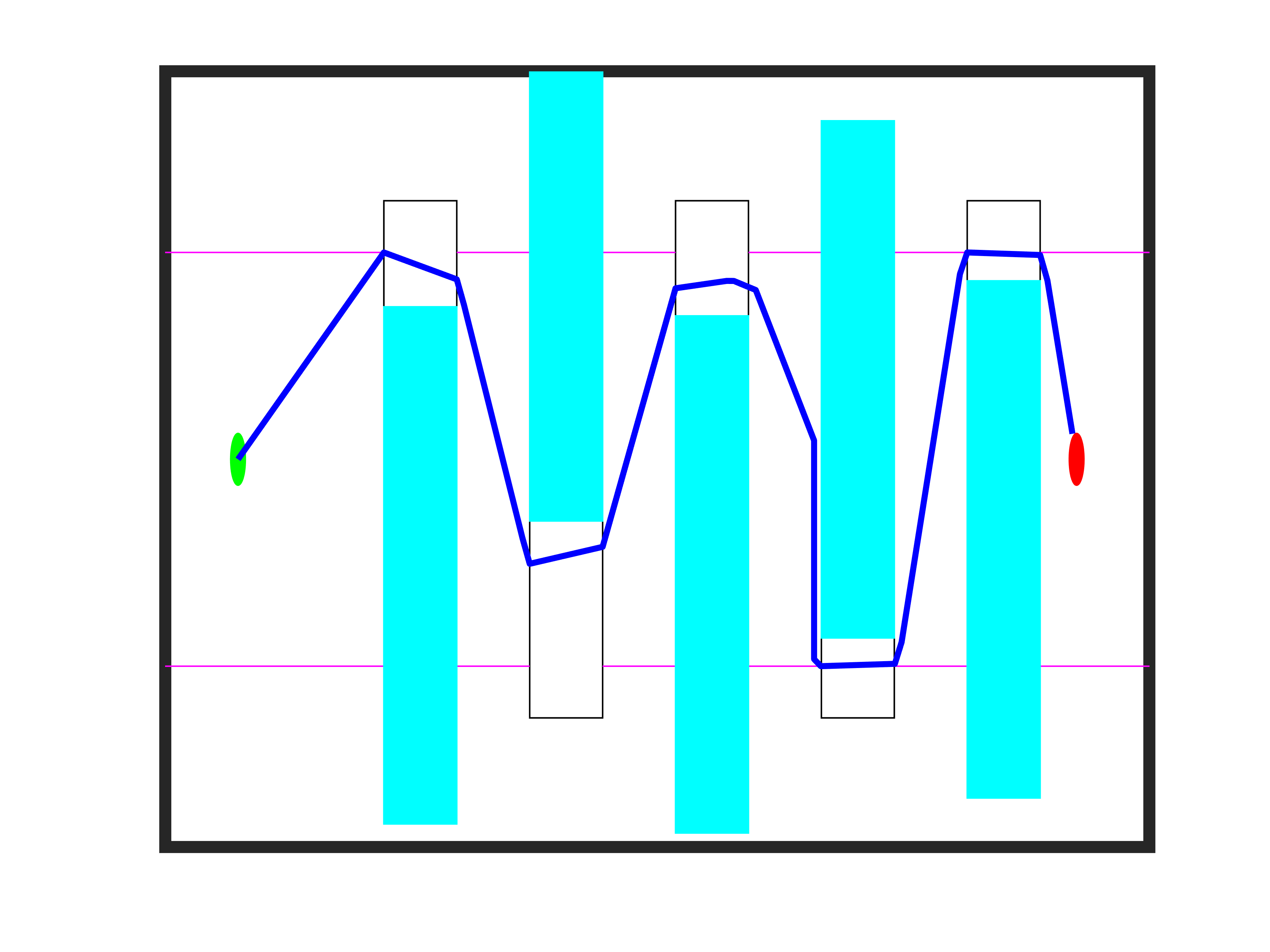}\label{fig:rec4}}
    \vspace{-0.2cm}
    \caption{A robot asked to move from left to right in the presence of 5 movable obstacles. The $y$-coordinate of the robot is constrained to lie within the magenta lines. (a) Obstacles can only translate in the vertical direction. $m=1$ and $m=2$ produced similar trajectories (obstacle displacements shown in cyan). (c)-(d) Odd obstacles (counting from left) can only move downward and even obstacles can only move upward. Optimal solution is given in (c). $m=2$ gives a sub-optimal solution as shown in (d) with the first sub-problem involving the first 3 obstacles (evident from the vertical trajectory after the third obstacle).}
  \label{fig:rectangle}
\end{figure*}
So far we have formulated the MOD problem as an optimization problem~\eqref{eq:optimization_problem}. In MIQP problems the worst-case computational complexity grows exponentially with the number of binary/integer decision variables~\cite{naik2018PHD}. We note here that for each obstacle we have introduced four binary variables to formulate the optimization as an MIQP problem. The number of binary variables is thus $2^{4 \times T \times |\mathcal{O}|}$, where $T$ is the number of time steps required to reach the goal. Thus, with increase in the number of movable obstacles and the environment size (larger the environment, larger is the $T$) the optimization tends to be computationally infeasible. The computational burden can be reduced by decreasing the number of variables. One way to achieve this is by solving subsets of the problem, each of which is computationally inexpensive compared to the solving the complete optimization problem at once. To reduce the computational burden (at the cost of optimality) we propose splitting the optimization problem into $m$ sub-problems, each of duration $T'$ such that $mT' = T$. We slice the number of time steps required to reach the goal into smaller duration and solve the optimization problem for each slice. Once each sub-problem is optimized, the values of time dependent variables are passed on to the following sub-problem. 

Let the optimal objective function without horizon slicing be denoted by $J^\star$. Let us also denote by $c_k$, the cost at each time step. $J^\star$ can thus be written as
\vspace{-0.3cm}
\begin{small}
\begin{equation}
\begin{split}
    J^\star &= \sum_{k=1}^T c^\star_k
     = \sum_{k=1}^{T'}c^\star_k + \ldots+ \sum_{k=(m-1)T'+1}^{T}c^\star_k\\
     &= J^\star_{1\rightarrow T'} + \ldots + J^\star_{(m-1)T'+1\rightarrow T}
     = \sum_{i=1}^m J^{\star}_{(i-1)T'+1\rightarrow iT'}
     \end{split}
     \label{eq:horizon_optimal}
\end{equation}
\end{small}
\noindent where $i \in \{1,\ldots,m\}$. Now let us consider the case where we slice the horizon and solve $m$ different optimization problems. The overall objective function $J^s$ is obtained by adding the individual objectives
\begin{equation}
    J^s = \sum_{k=1}^T c_k = \sum_{i=1}^m J^s_{(i-1)T'+1\rightarrow iT'}
    \label{eq:horizon_sliced}
\end{equation}
\noindent where $J^s_{(i-1)T'+1\rightarrow iT'}$ represent the objective function of the $i-$the sub-problem. As noted before, from an algorithmic perspective, a subset of an optimal solution is not necessarily an optimal solution. However, it is easily verified that the solutions to all the subsets is naturally a feasible solution. Thus, the  horizon slicing approach results in a feasible solution to the original problem and therefore the value of $J^s$ must be an upper bound, that is, $J^s \geq J^\star$. Since $J^s$ is an upper bound, we have
\begin{small}
\begin{equation}
\begin{split}
    J^s & = \sum_{i=1}^m J^s_{(i-1)T'+1\rightarrow iT'}
      = \sum_{i=1}^m \left( J^{\star}_{(i-1)T'+1\rightarrow iT'} + \delta^i \right)\\
     & \leq \left( \sum_{i=1}^m  J^{\star}_{(i-1)T'+1\rightarrow iT'} \right) + m\delta^{max}
     \leq J^\star + m\delta^{max}
\end{split}
\label{eq:upperbound_J}
\end{equation}
\end{small}
noindent where $\delta^i = J^s_{(i-1)T'+1\rightarrow iT'} - J^{\star}_{(i-1)T'+1\rightarrow iT'}$ and $\delta^{max} = \max \{\delta^i \ | 1 \leq i \leq m \} $. From~\eqref{eq:upperbound_J}, we thus have
\begin{equation}
    J^\star \leq J^s \leq J^\star + m\delta^{max}
    \label{eq:epsilon_boud}
\end{equation}
Let the deviation from the optimal solution $J^s - J^\star$ be a factor of $J^\star$, that is, $J^s-J^\star = \epsilon J^\star$. Ideally, we would want $\epsilon \to 0$ with considerable computational efficiency. In the optimization literature the standard metric used to compute the relative distance between an approximate solution and the optimal solution is the \textit{optimality gap}. $\epsilon$ essentially computes this relative distance as shown below
\begin{equation}
   \textrm{optimality gap} = \frac{J^s- J^\star}{J^\star} = \frac{\epsilon J^\star}{J^\star} = \epsilon
\end{equation}
From~\eqref{eq:epsilon_boud}, we have $J^s- J^\star \leq m \delta^{max} \implies \epsilon J^\star \leq m \delta^{max} $ and therefore $\epsilon \leq m \delta^{max}/J^{\star}$. Clearly, for a given problem as the number of horizon slices or sub-problems increase, the value of $\epsilon$ also grows. The exact value is problem specific (see Fig.~\ref{fig:rectangle}) since it depends on a number of parameters such as the robot and obstacle dynamics, size of the environment, distribution of the obstacles in the environment. Yet, it is easily verified that horizon slicing computes a $m-$approximation as $\delta_{max}/J^\star < 1$ since $\delta^{max}$ is the maximum deviation of a sub-problem from its corresponding optimal solution, giving $\epsilon < m $. In Section~\ref{sec:results}, we provide an empirical analysis to determine the best $m$ by performing different experiments while varying some of the parameters stated above.

%% file: results.tex
In this section we evaluate our approach using (1) the naive approach that is computationally expensive and (2) horizon slicing. The Optimization problem in~\eqref{eq:optimization_problem} is performed using the IBM ILOG CPLEX Optimization Studio V12.10.0 in MATLAB under YALMIP interface \cite{Lofberg2004}. Our method is tested on three different domains, namely, (1) \textit{SQUARE} domain with 36 obstacles, (2) \textit{ROBOT} domain with 74 obstacles, and (3) \textit{RANDOM} domain with 100 obstacles. As discussed in Section~\ref{sec:approach}, $m$ denotes the number of horizon slices and $m=1$ correspond to the naive approach without the horizon slicing. Currently, the values of $\alpha^{re}$, $\alpha^r$ and $\alpha^o$ are chosen via empirical tuning and for all the experiments we use $\alpha^{re} = 10$, $\alpha^r = 0.5$ and $\alpha^o= 100$.  The performance is evaluated on an Intel{\small\textregistered} i7-10850H CPU @ 2.70GHz with 32 GB RAM under Windows 11.  
\begin{figure}[t!]
   \subfloat[\textit{SQUARE}]{\includegraphics[trim=5 1 5 1,clip,scale=0.41]{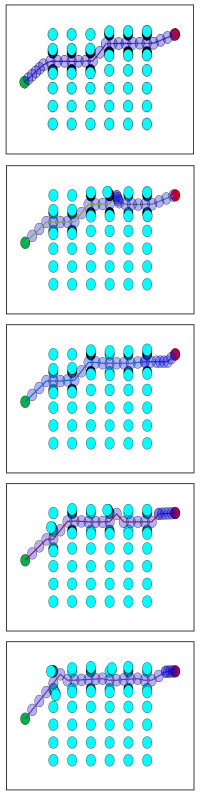}\label{fig:re1}}\hfill
    \subfloat[\textit{ROBOT}]{\includegraphics[trim=5 1 5 1,clip,scale=0.41]{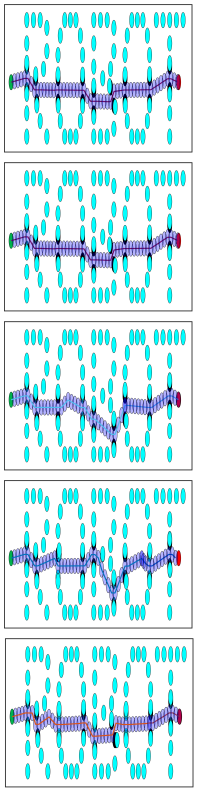}\label{fig:re2}}\hfill
  \subfloat[\textit{RANDOM}]{\includegraphics[trim=5 1 5 1,clip,scale=0.41]{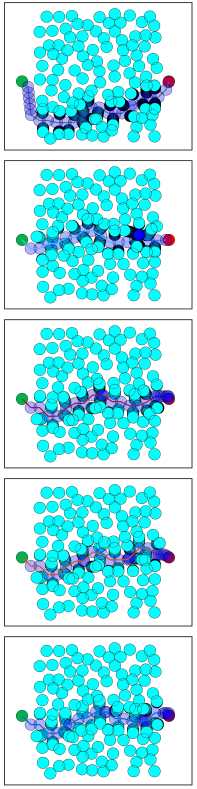}\label{fig:re3}}
      \label{fig:re4}
      \vspace{-0.2cm}
       \caption{Robot trajectory and obstacle displacements while the robot moves from left to right under model~\eqref{eq:robot_model1}. Figures correspond to $m=1,\ldots,5$ starting from the first row. }
  \label{fig:linearmodel1}
\end{figure}
\begin{figure}[t!]
    \subfloat{\includegraphics[trim=440 220 320 100,clip,scale=0.028]{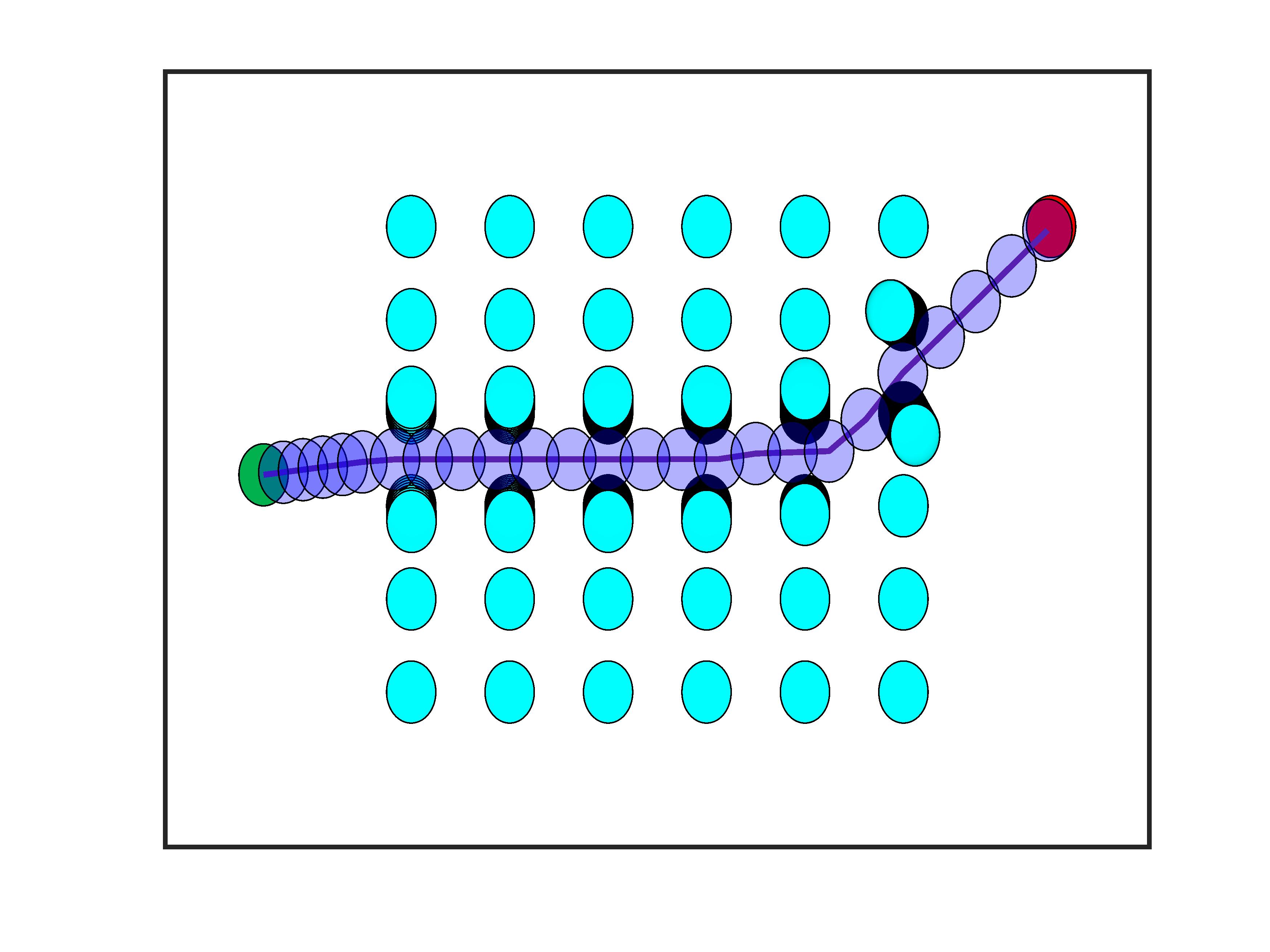}\label{fig:C61}}\hfill
     \subfloat{\includegraphics[trim=440 220 320 100,clip,scale=0.028]{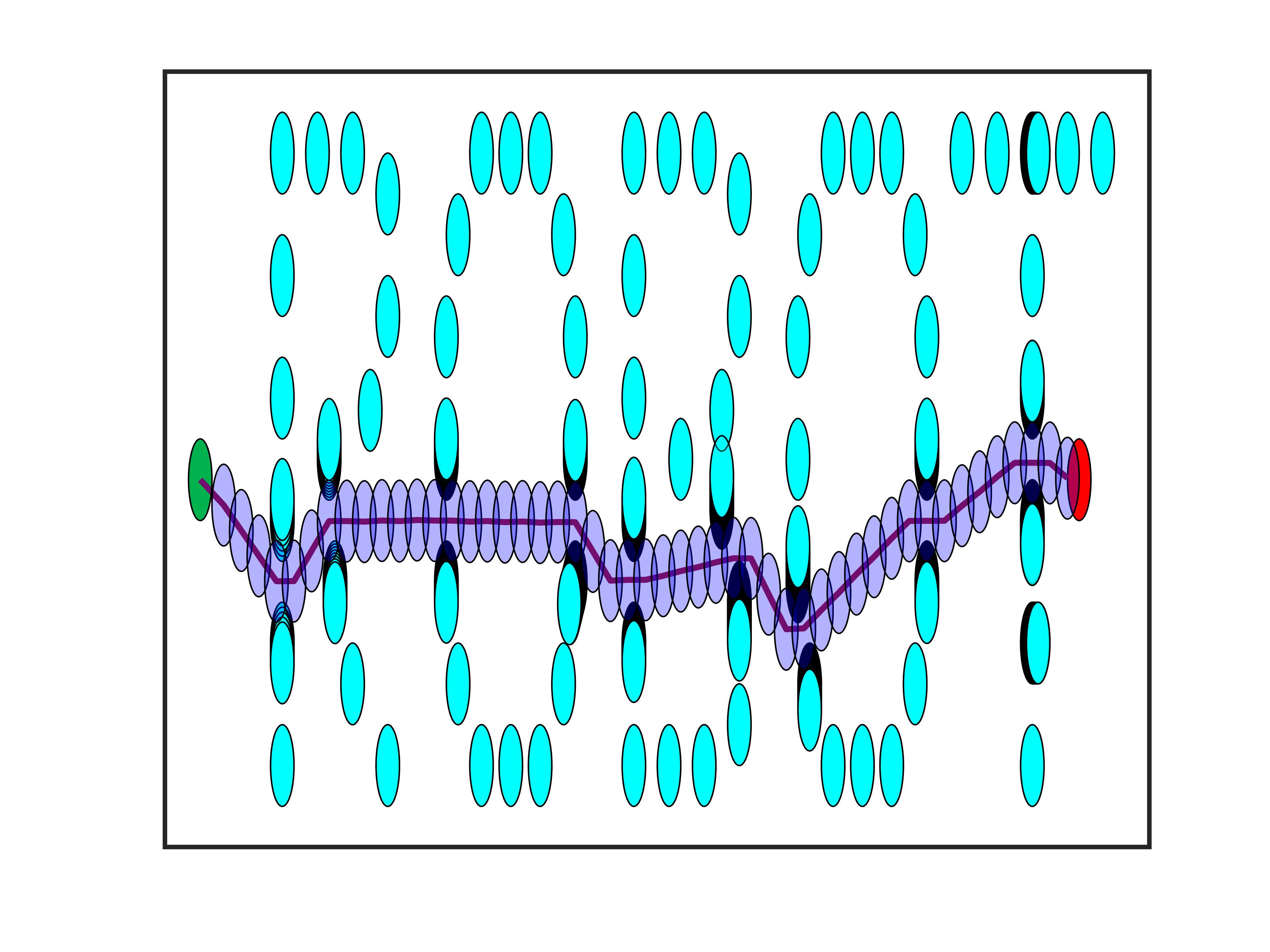}\label{fig:C71}}\hfill
 \subfloat{\includegraphics[trim=440 220 320 100,clip,scale=0.028]{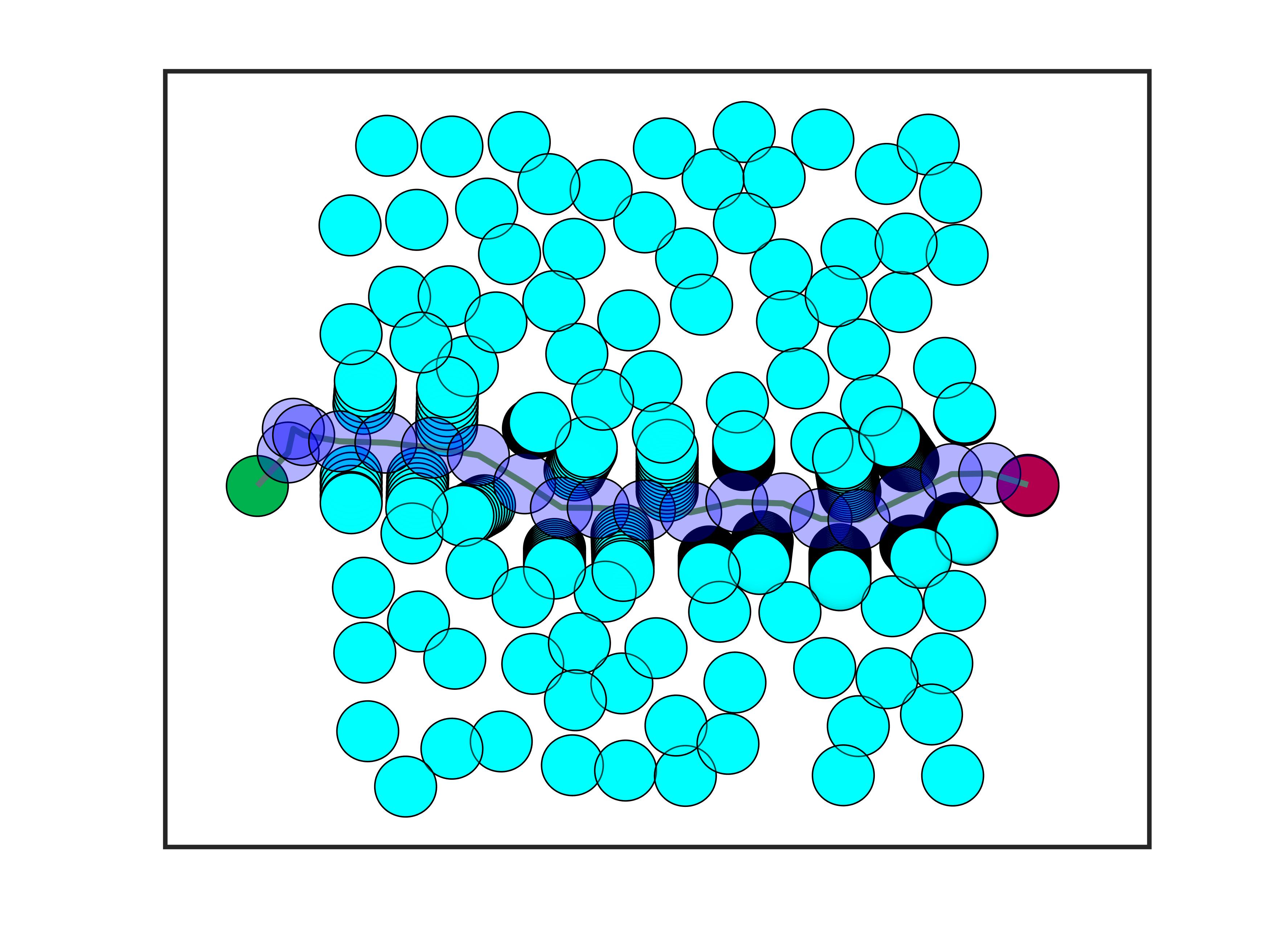}\label{fig:C81}}\hfill\\
 
    \subfloat{\includegraphics[trim=440 220 320 100,clip,scale=0.028]{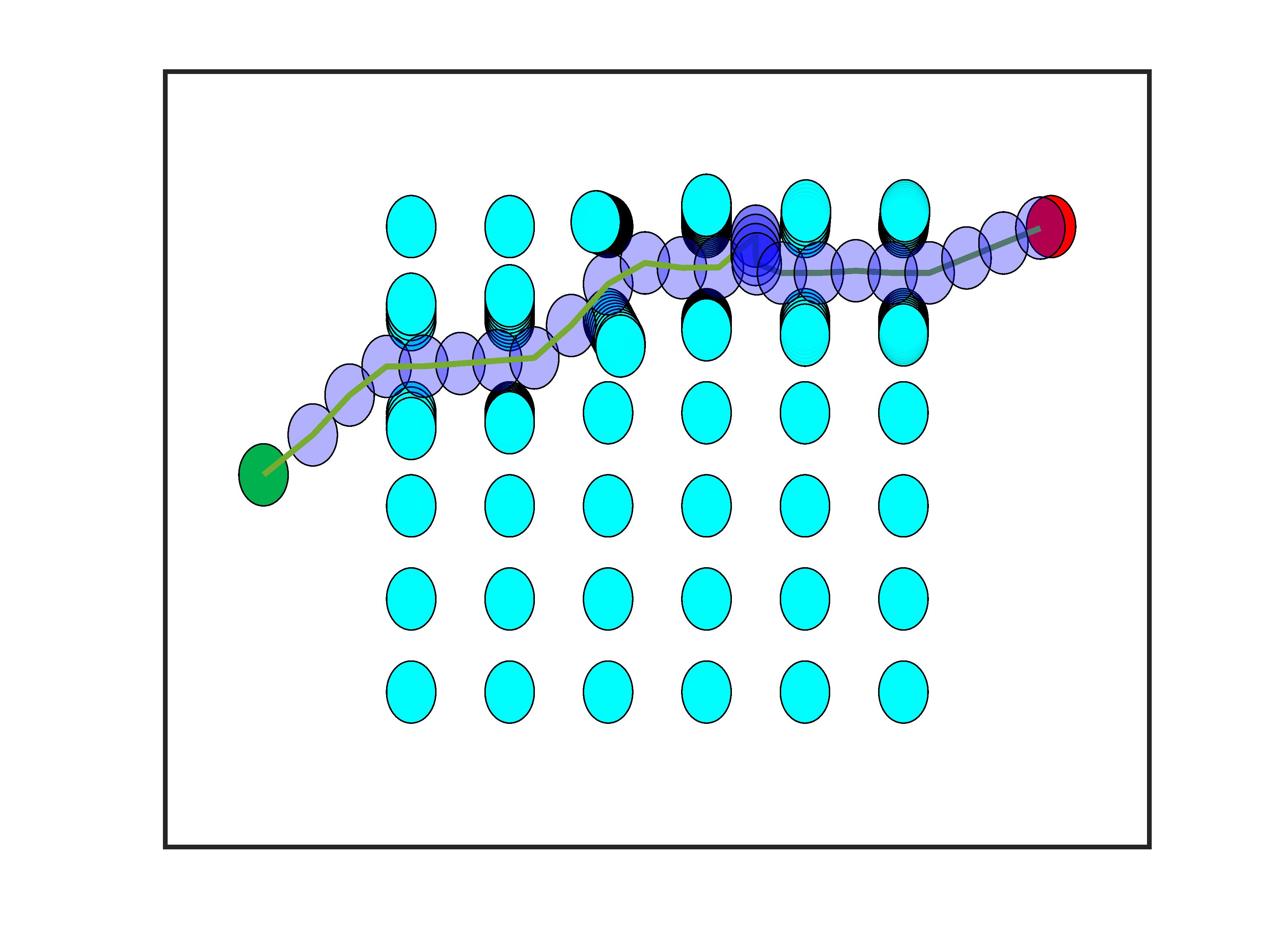}\label{fig:C62}}\hfill
   \subfloat{\includegraphics[trim=440 220 320 100,clip,scale=0.028]{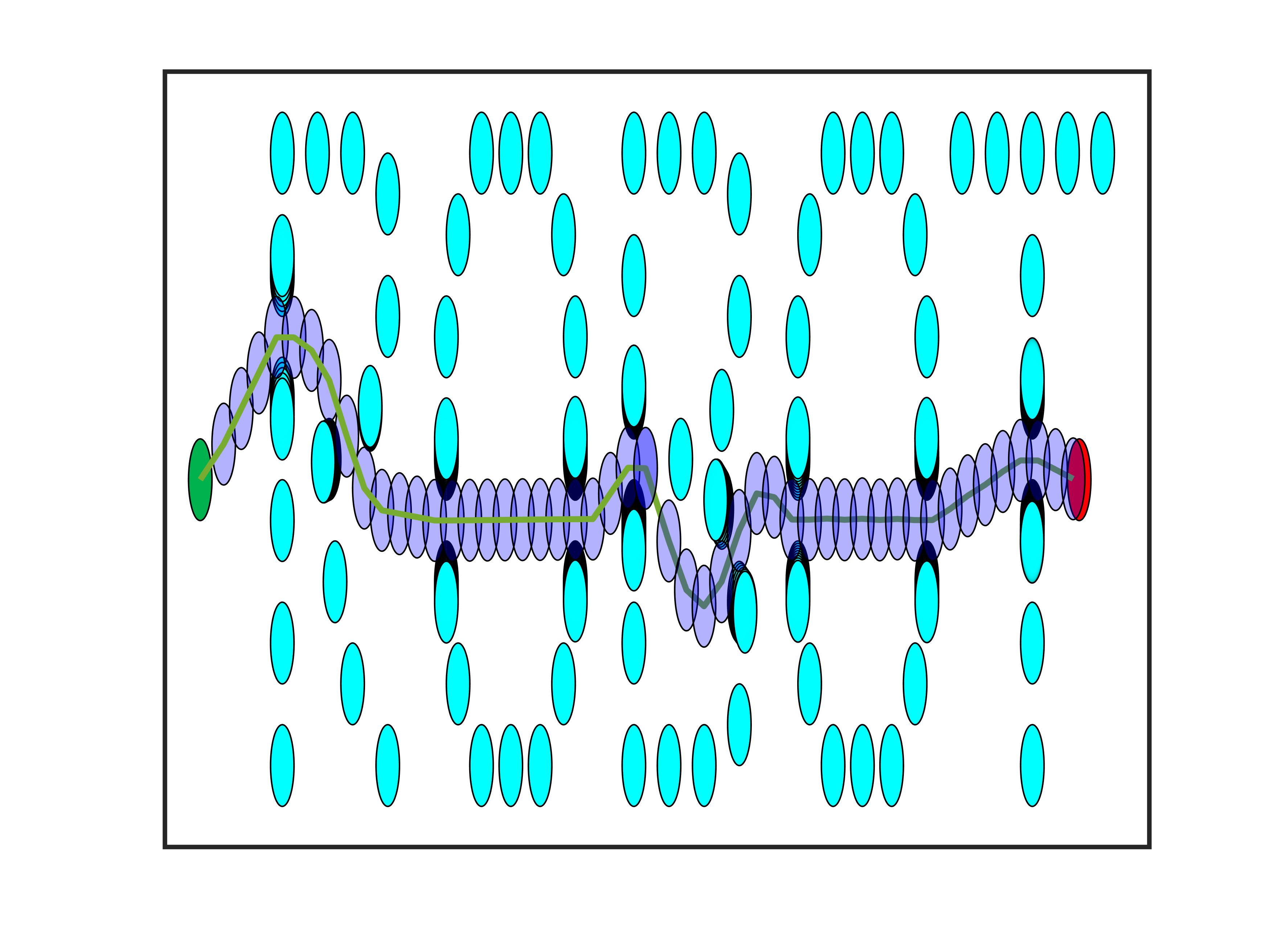}\label{fig:C72}}\hfill
   \subfloat{\includegraphics[trim=440 220 320 100,clip,scale=0.028]{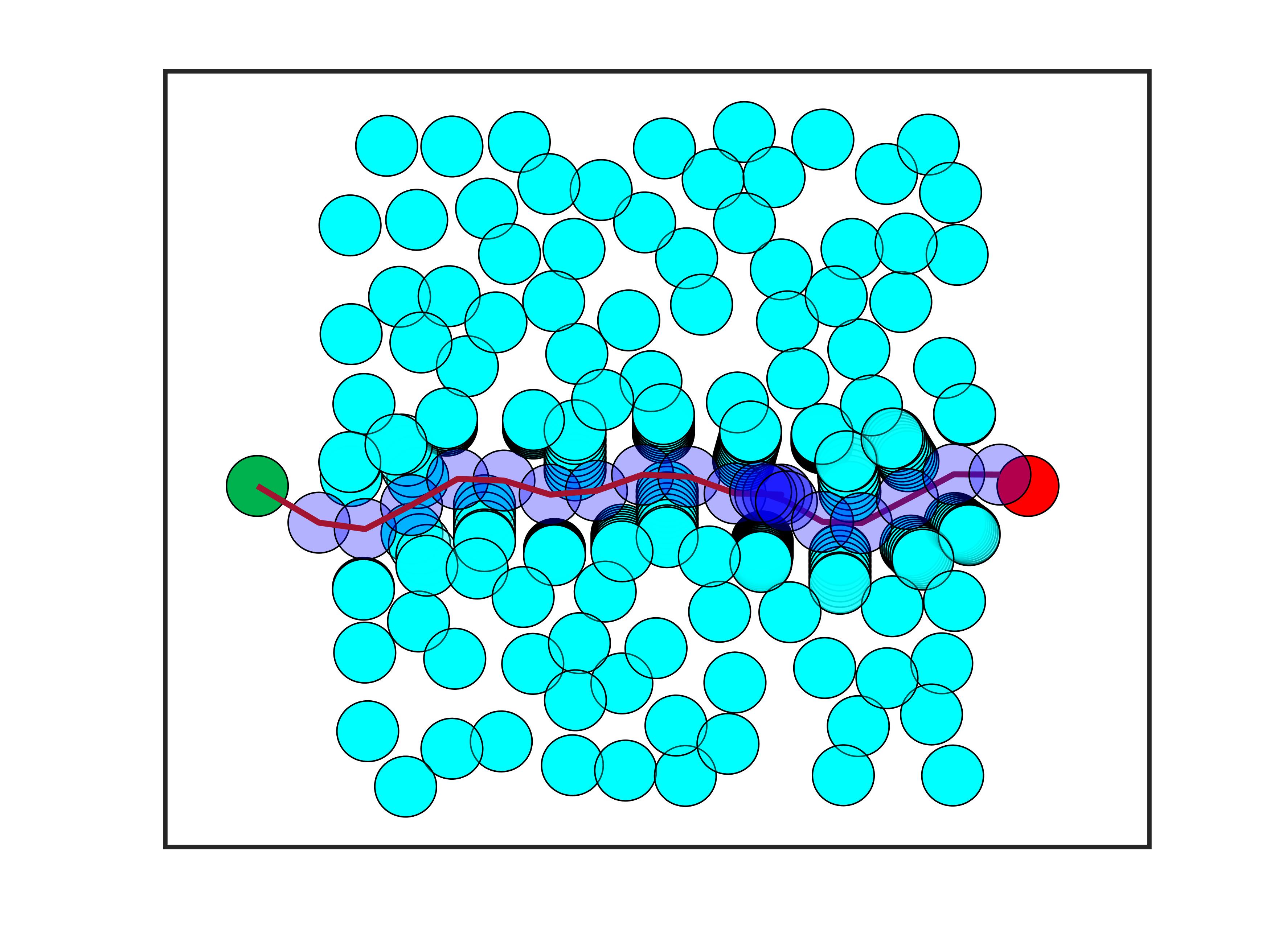}\label{fig:C82}}\hfill\\
   
  \subfloat{\includegraphics[trim=440 220 320 100,clip,scale=0.028]{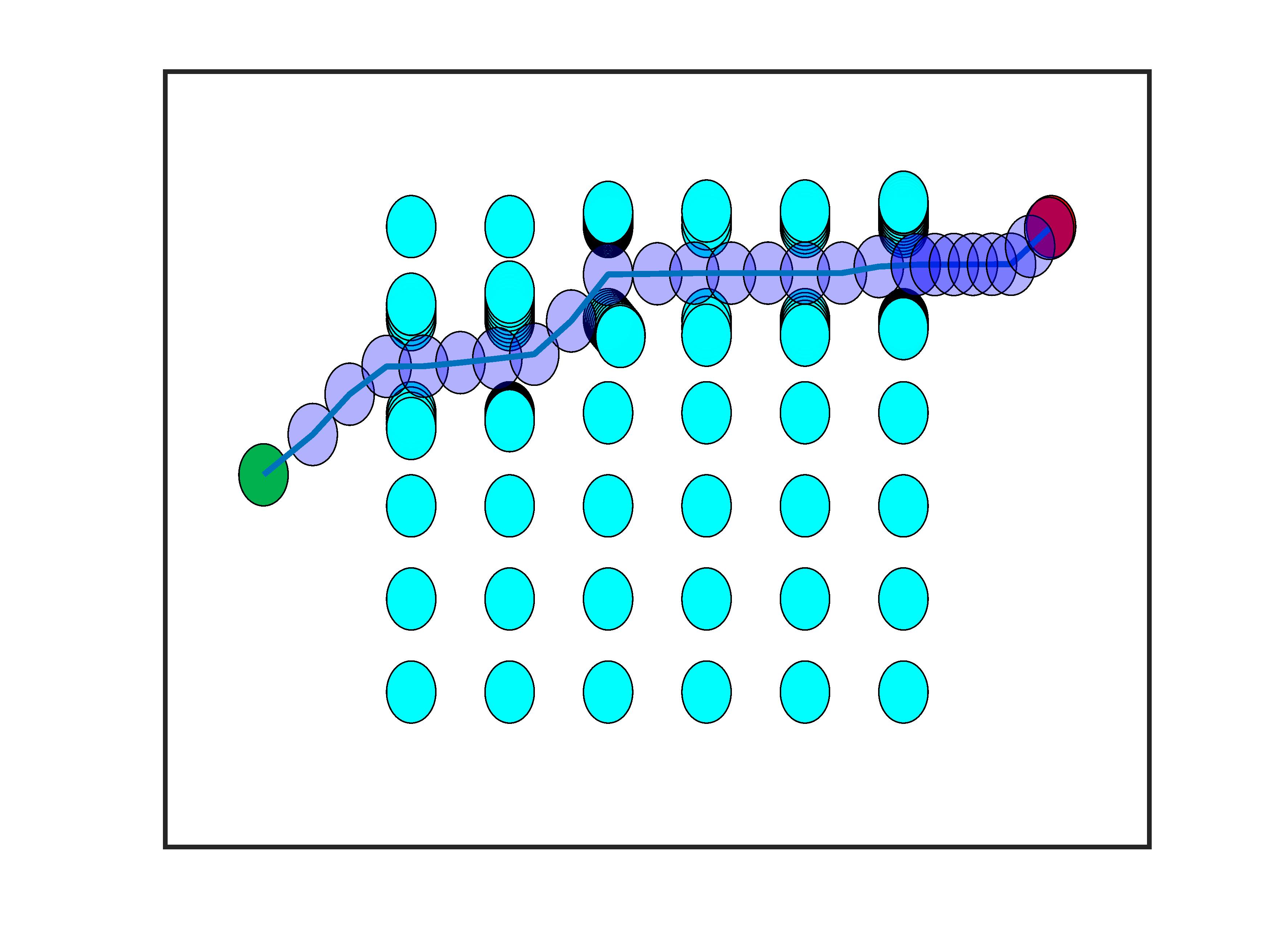}\label{fig:C63}}\hfill
  \subfloat{\includegraphics[trim=440 220 320 100,clip,scale=0.028]{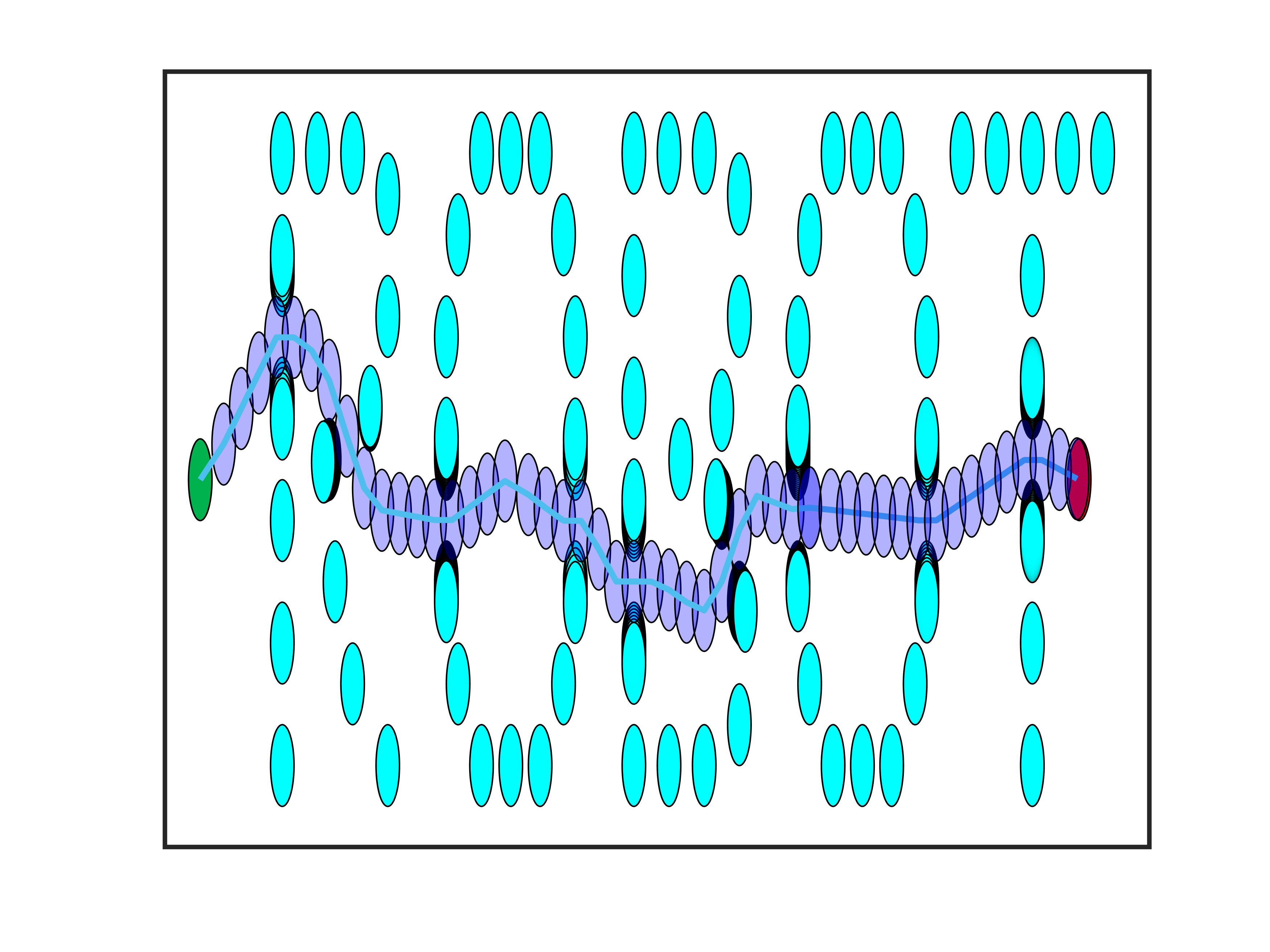}\label{fig:C73}}\hfill
  \subfloat{\includegraphics[trim=440 220 320 100,clip,scale=0.028]{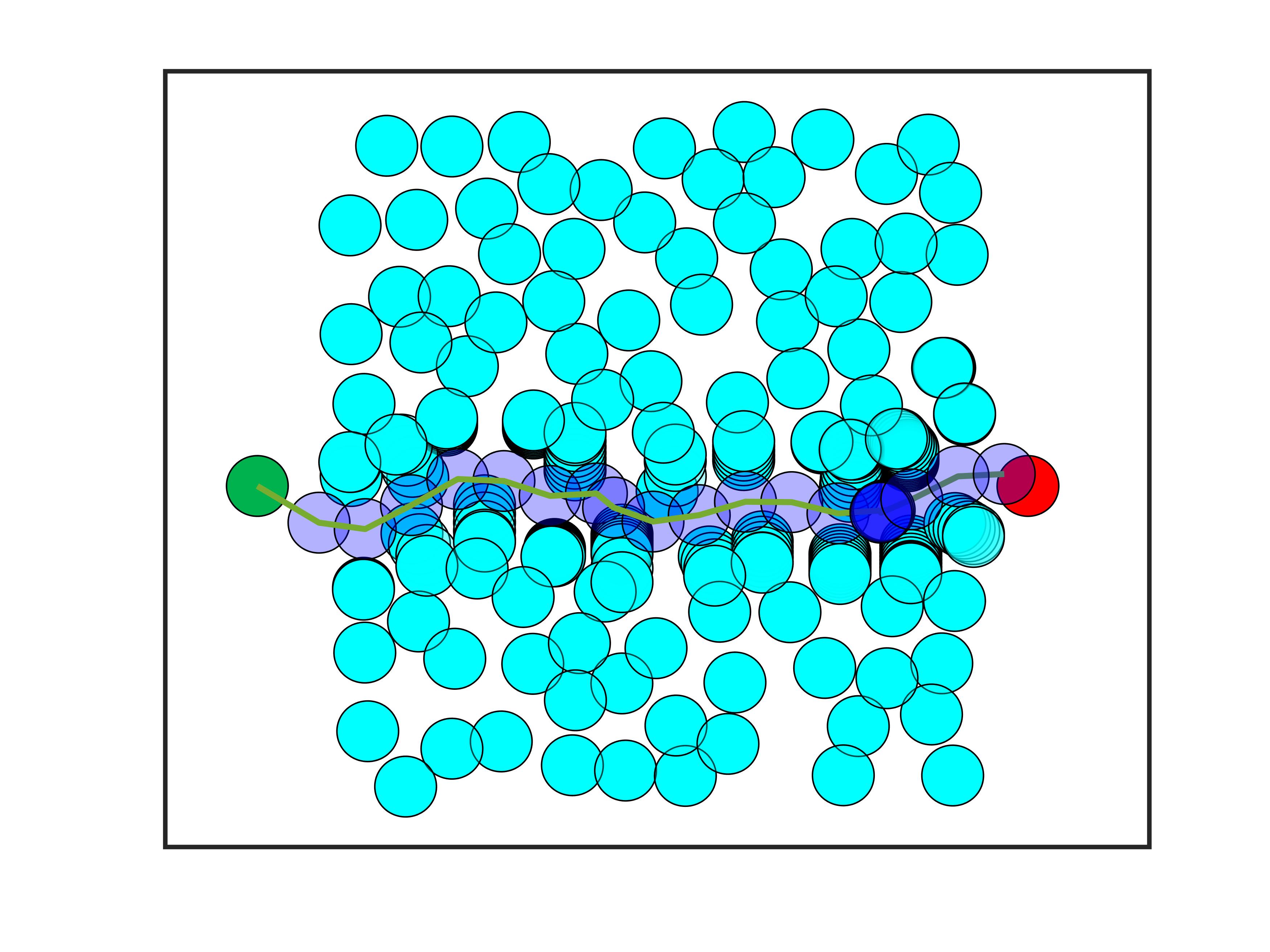}\label{fig:C83}}\hfill\\
  
   \subfloat{\includegraphics[trim=440 220 320 100,clip,scale=0.028]{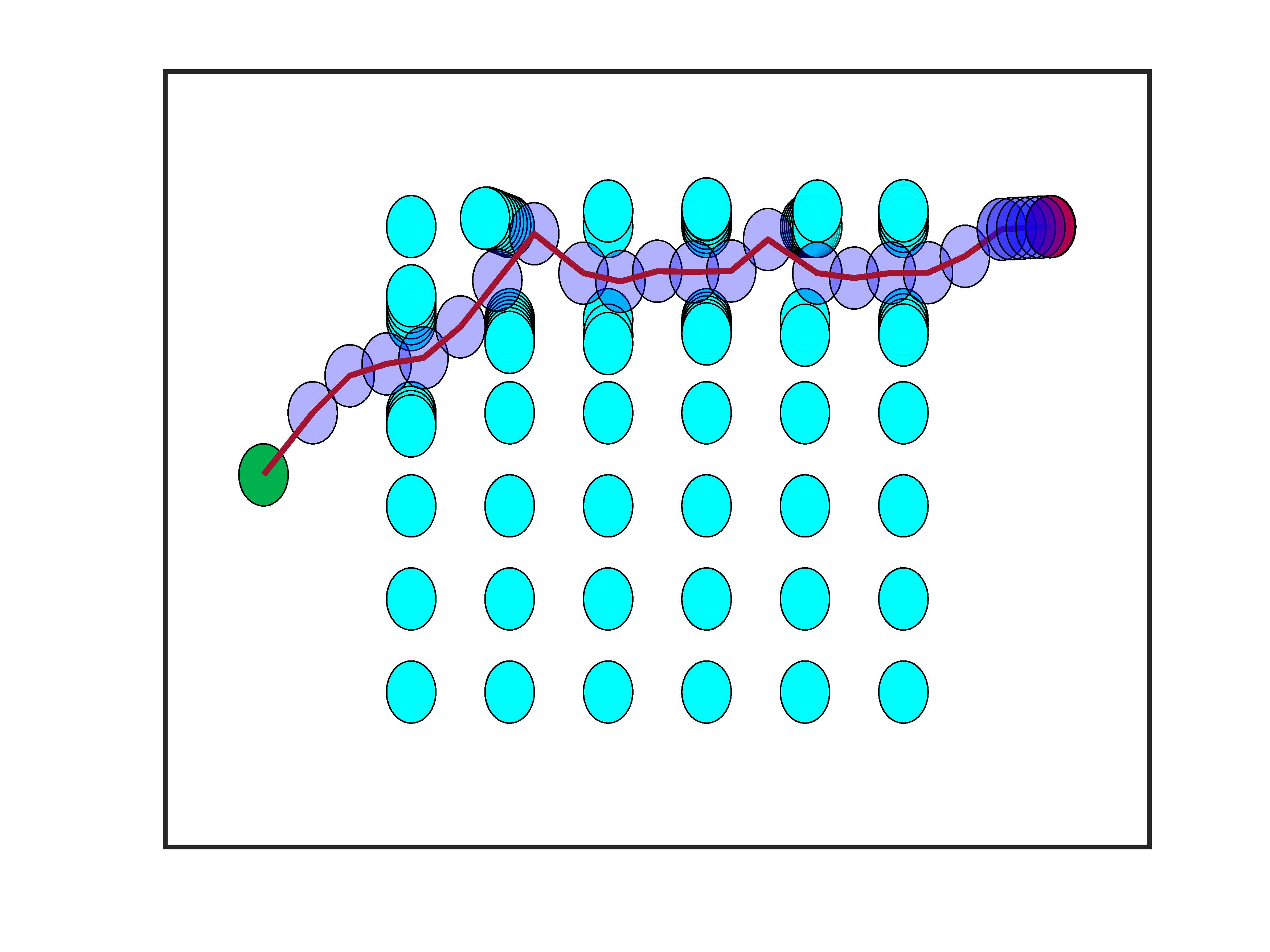}\label{fig:C64}}\hfill
   \subfloat{\includegraphics[trim=440 220 320 100,clip,scale=0.028]{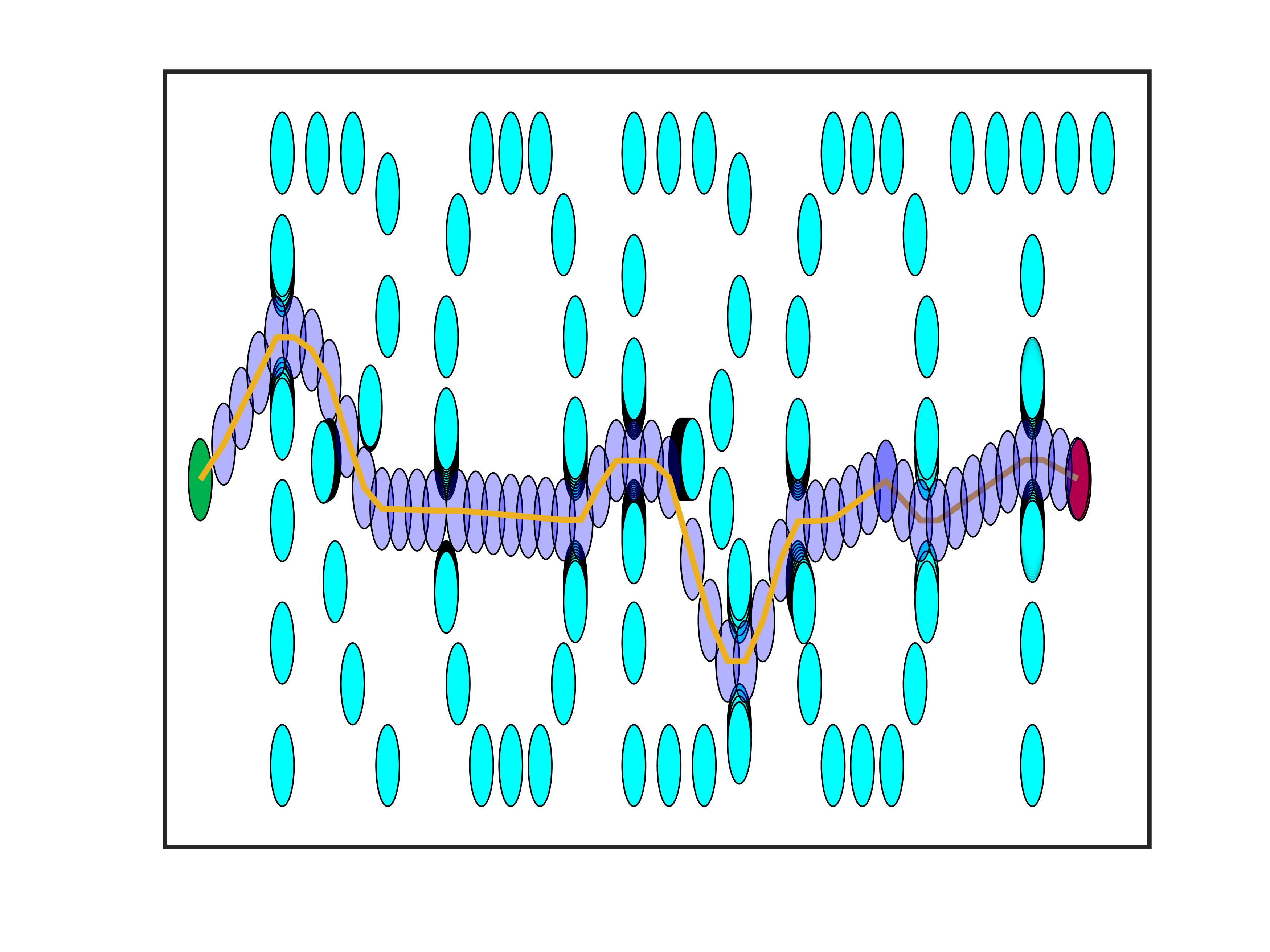}\label{fig:C74}}\hfill
   \subfloat{\includegraphics[trim=440 220 320 100,clip,scale=0.028]{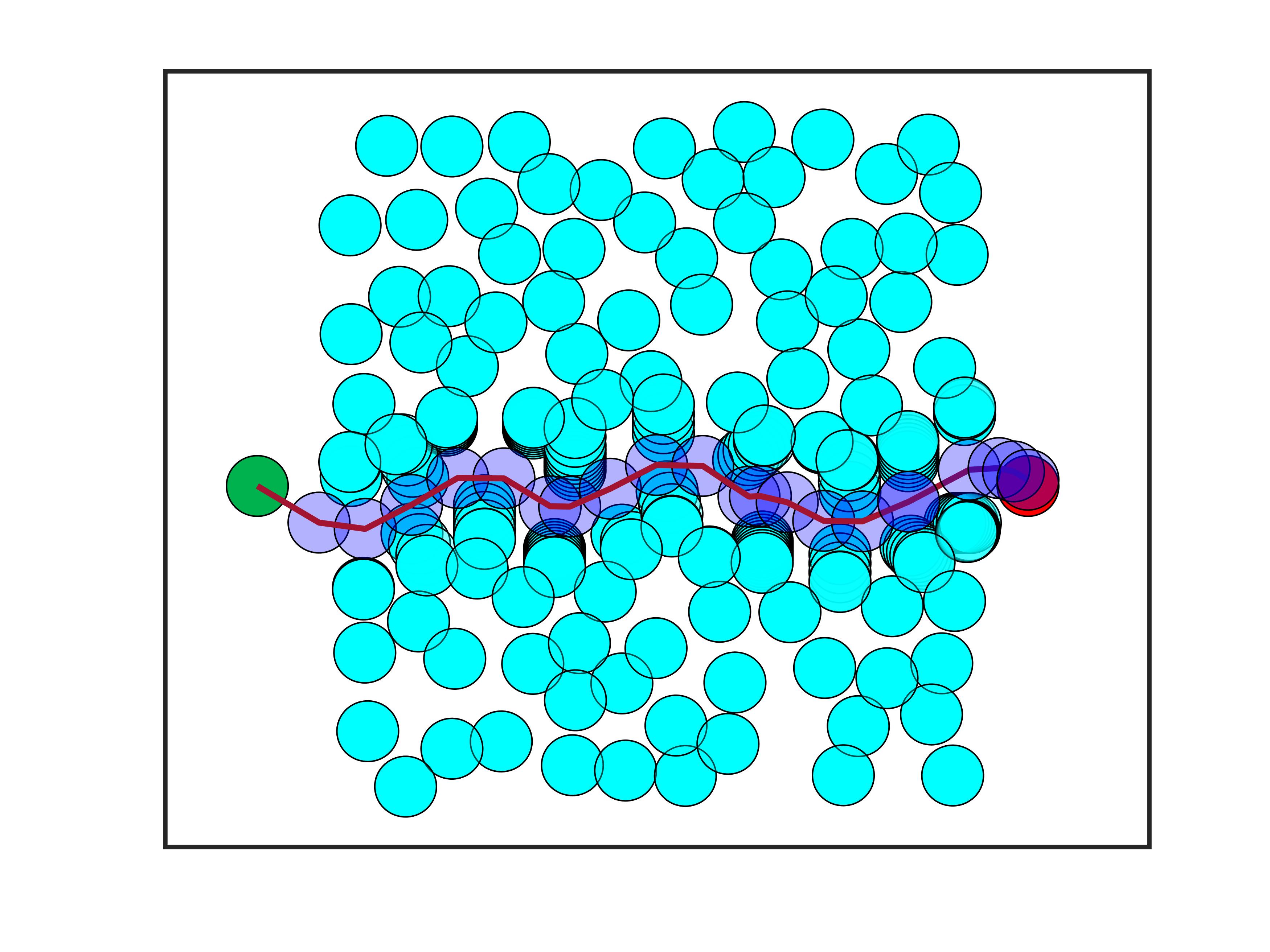}\label{fig:C84}}\hfill\\
   
   \clearsubcaptcounter
     \subfloat[\textit{SQUARE}]{\includegraphics[trim=440 220 320 100,clip,scale=0.028]{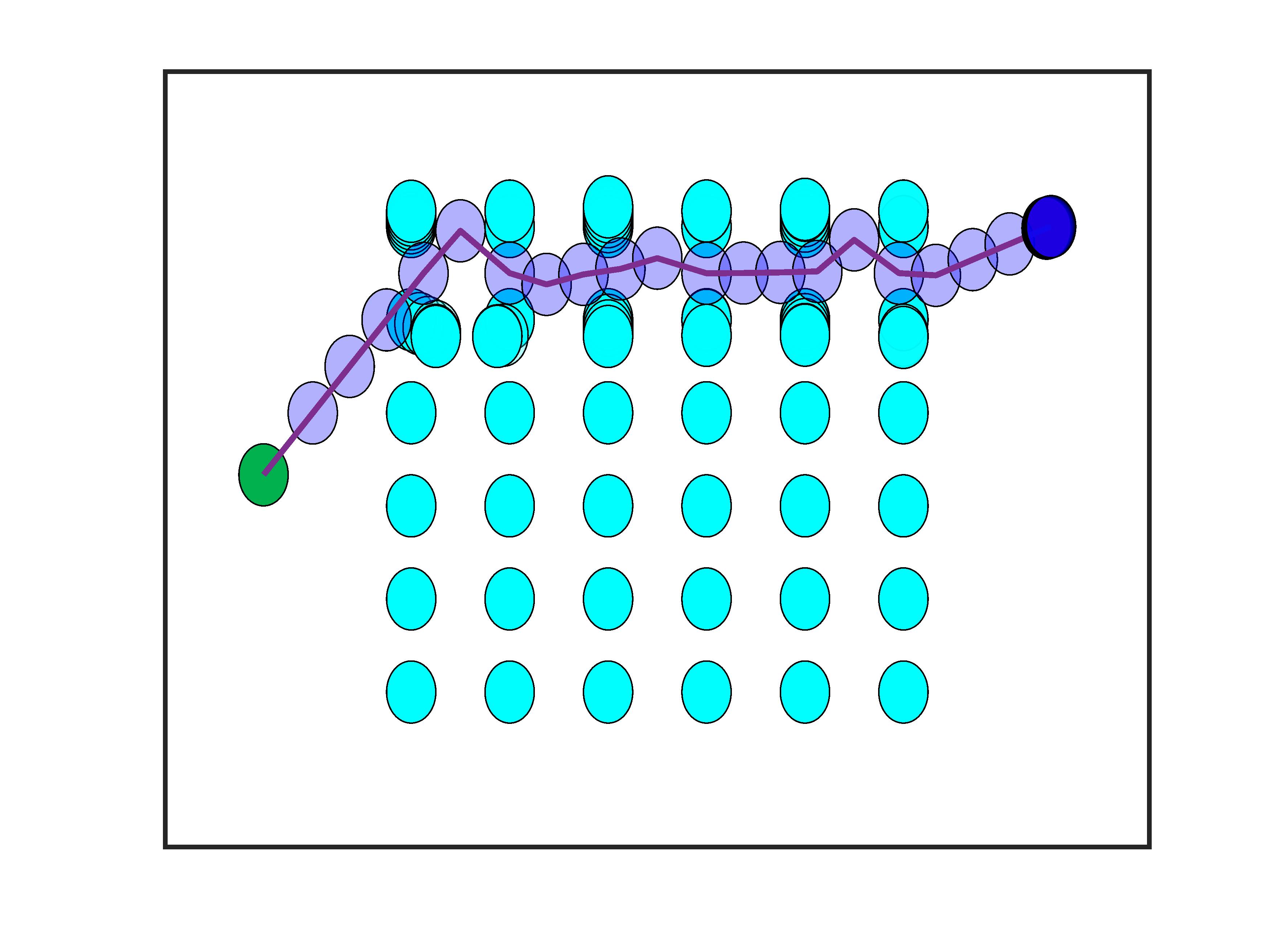}\label{fig:C65}}\hfill   \subfloat[\textit{ROBOT}]{\includegraphics[trim=440 220 320 100,clip,scale=0.028]{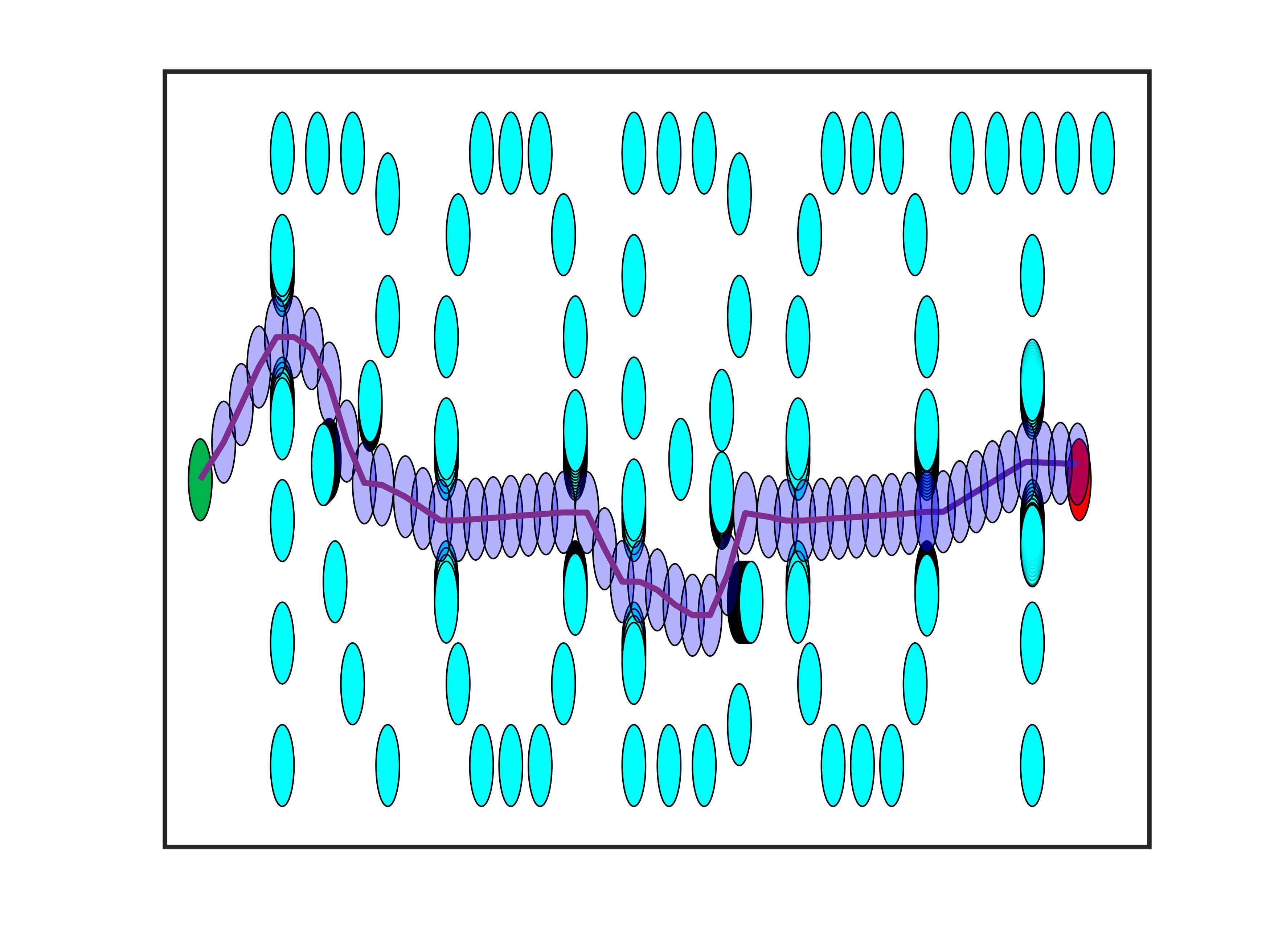}\label{fig:C75}}\hfill
   \subfloat[\textit{RANDOM}]{\includegraphics[trim=440 220 320 100,clip,scale=0.028]{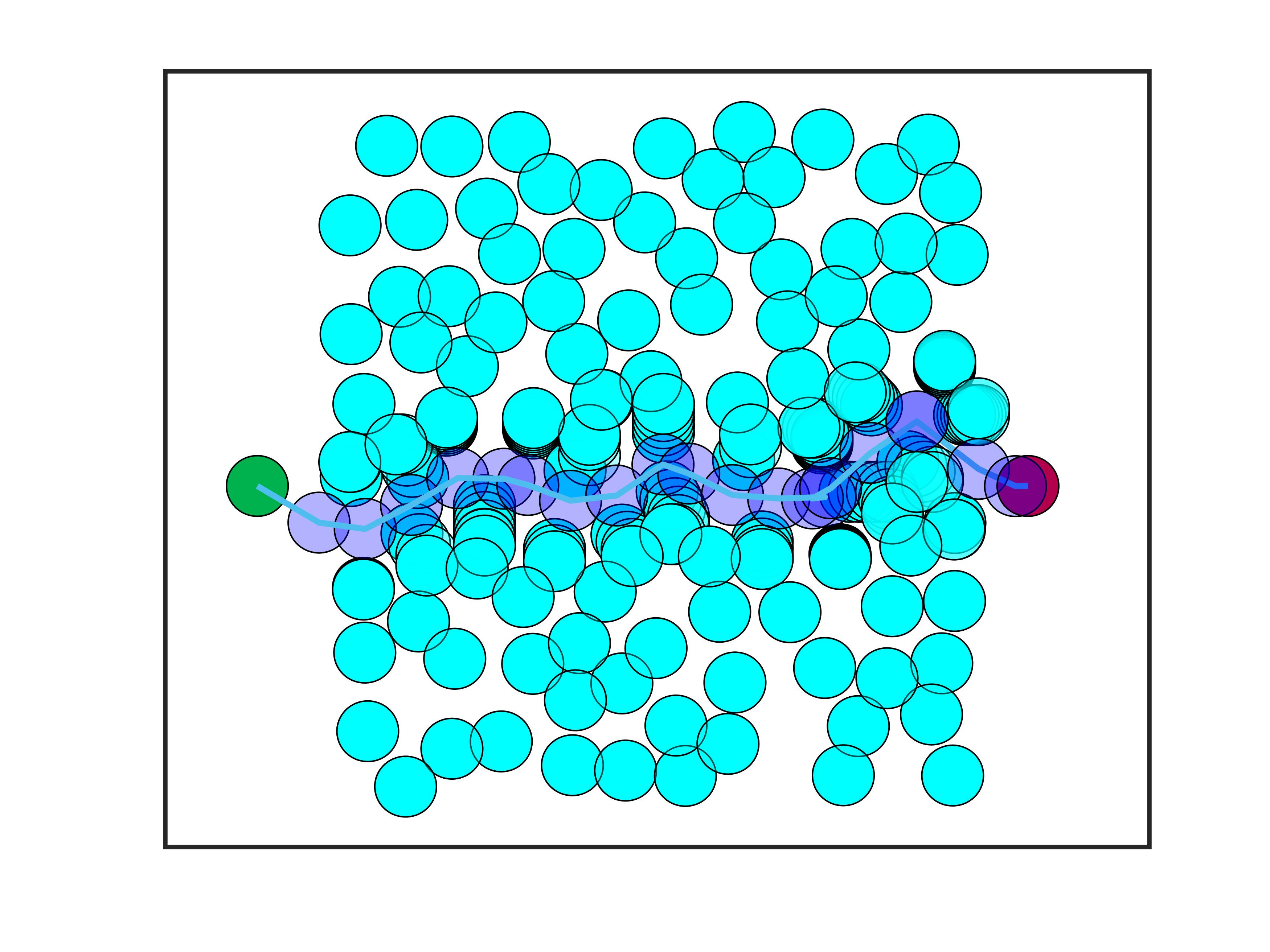}\label{fig:C85}}\hfill
   \vspace{-0.2cm}
    \caption{Robot trajectory and obstacle displacements while employing the robot model in~\eqref{eq:robot_model2}. Figures correspond to $m=1,\ldots,5$ starting from the first row.} 
  \label{fig:linearmodel2}
\end{figure}


We first test our method employing the robot dynamics given in~\eqref{eq:robot_model1}. The robot trajectory and the obstacle displacements computed are visualized in Fig.~\ref{fig:linearmodel1}. Various statistics for the same are presented in Table~\ref{table:result1} under the Robot model~\eqref{eq:robot_model1} column. The naive optimization (without horizon slicing) is denoted by $m=1$ and is computationally expensive as seen from Table~\ref{table:result1}. As it has been argued before, the computational complexity is exponential in the number of binary variables (proportional to the number of movable obstacles) and is evident from the computational time in \textit{ROBOT} and \textit{RANDOM} domains, with 74 and 100 obstacles, respectively. Additionally, the complexity also depends on (1) the spread of the obstacles with respect to the start and goal location of the robot and (2) the size of the environment. These two factors result in higher computation time for the \textit{ROBOT} domain ($79$ m$\times18$ m) with 74 obstacles than the \textit{RANDOM} domain ($24$ m$\times24$ m) with 100 obstacles. Analogous results are seen when employing the linear robot model in~\eqref{eq:robot_model2}--- the robot trajectory and obstacle movements visualized in Fig.~\ref{fig:linearmodel2}. Clearly, in all the domains, for both the linear robot models, as $m$ increases, the deviation from the optimal solution increases resulting in sub-optimal solutions as corroborated by the rise in $\epsilon$ values. Yet, higher values of $m$ results in great computational efficiency. Keeping in mind a pragmatic tradeoff between optimality and computational efficiency, the statistics in Table~\ref{table:result1} suggest that for small number of movable obstacles, $m=2$ may be used and for a larger number, $m=3$ may be employed.
  \setlength{\tabcolsep}{4pt}
	\begin{table}[]
	\scalebox{0.65}
		{       
							\begin{tabular}{l l rccc || rccc }        
							\hline 	
					\multicolumn{1}{c}{\multirow{ 1}{*}{}} &  \multicolumn{1}{c}{$m$} &  \multicolumn{4}{c}{Robot model~\eqref{eq:robot_model1}} & \multicolumn{4}{c}{Robot model~\eqref{eq:robot_model2}}\\ [0.5ex]
					\cline{3-6}\cline{7-10} \\ [-8pt]
																{} & {} & \multicolumn{1}{c}{CPU [$s$]} & \multicolumn{1}{c}{Gap/$\epsilon$}  & \multicolumn{1}{c}{$J^r [m^2]$} &  \multicolumn{1}{c}{$J^o [m^2]$} & \multicolumn{1}{c}{CPU [$s$]} & \multicolumn{1}{c}{Gap/$\epsilon$}  & \multicolumn{1}{c}{$J^r [m^2]$} &  \multicolumn{1}{c}{$J^o [m^2]$} \\
				 [0.5ex]
		       \hline    
		       
	 &	1 & 1601.58   & 0.00      & 51.20  & 0.28 & 1000.98   & 0.00      & 51.22  & 0.28 \\ 
&2 & 27.92 & 0.62 & 59.17  & 0.56 & 116.37 & 0.76 & 59.33  & 0.65 \\ 
\textit{\textit{SQUARE}} &3 & 4.79    & 0.90 & 53.87  & 0.74 & 7.19    & 1.06 & 54.38  & 0.84 \\ 
&4 & 1.32   & 2.10   & 59.44  & 1.35& 1.95   & 3.03   & 64.91  & 1.85 \\
&5 & 1.02   & 3.08   & 59.29 & 1.88& 1.36   & 3.72   & 65.90 & 2.22 \\                 
                     \hline 
                     \\ [-8pt]
                     
		       \hline    
		       
	  &   1 & 28814.74   & 0.00     & 122.99  & 0.28 & 28808.67   & 0.00     & 132.70  & 0.32  \\ 
&2 & 17367.71 & 0.25 & 121.64  & 0.52  & 19564.10 & 0.18 & 133.00  & 0.49  \\ 
\textit{\textit{ROBOT}}	& 3 & 128.68    & 0.37 & 121.08 & 0.63 & 147.85    & 0.31 & 127.93 & 0.66  \\
&4 & 24.23   & 0.98   & 119.32  & 1.19  & 28.83   & 0.71   & 136.78 & 0.99 \\ 
&5& 11.62   & 1.21   & 142.74  & 1.28 & 12.24   & 0.75   & 128.16  & 1.08 \\               
                     \hline 
                     \\ [-8pt]
		       \hline    
		       
	  &   1 & 16009.18   & 0.00     & 51.86  & 1.13 & 18010.40   & 0.00     & 39.15  & 1.60 \\ 
&2 & 12813.12 & 0.78 & 49.56  & 2.24  & 14413.58 & 0.77 & 52.28 & 2.93   \\ 
\textit{\textit{RANDOM}}	& 3 & 6403.36   & 0.99 & 46.68 & 2.55  & 771.53    & 1.35 & 48.39  & 3.99 \\
&4 & 897.14   & 1.83   & 48.85  & 3.71 & 44.76   & 1.43  & 43.72  & 4.16 \\ 
&5& 14.43   & 2.05   & 46.70& 4.03 & 20.47   & 2.16   & 47.55 & 5.44\\  
                     \hline
		\end{tabular}  }      
		\vspace{-0.2cm}                            
		\caption{Different statistics for the three domains. CPU denotes the computation time in seconds for the optimization problem in~\eqref{eq:optimization_problem}. For $m>1$, $J^r$, $J^o$ represent the added cost across each slice.} 
\label{table:result1}
	\end{table}

%% file: discussion.tex
This paper looked into the MOD problem from a mobile robot navigation perspective. In addition to an exact method, a horizon slicing approach is presented which computes an $m-$approximate solution with high computational gain.

\begin{figure}[]
    \centering
    \scalebox{0.72}{
    \begin{tabular}{cc}
    \adjustbox{valign=b}{\subfloat[]{%
          \includegraphics[trim=0 0 50 0,clip,scale=0.4]{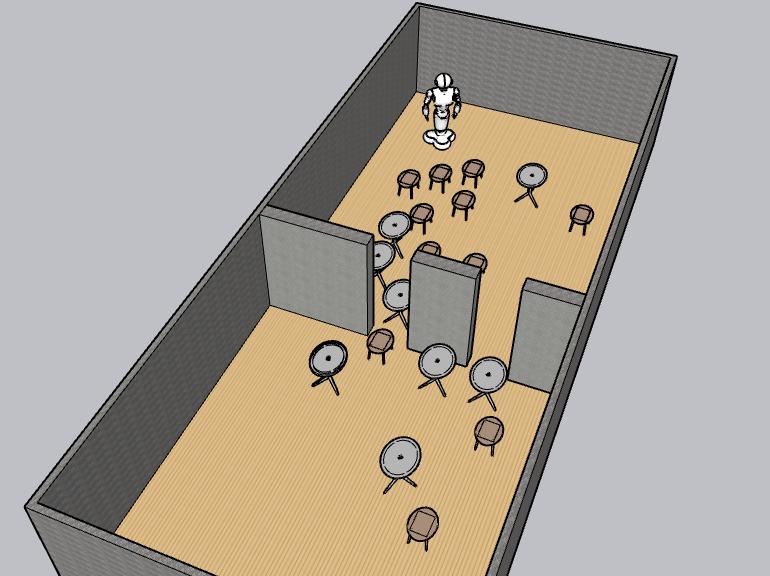}\label{fig:A1}}}
    &      
    \adjustbox{valign=b}{\begin{tabular}{@{}c@{}}
    \subfloat[]{%
          \includegraphics[trim=50 35 30 10,clip,scale=0.3]{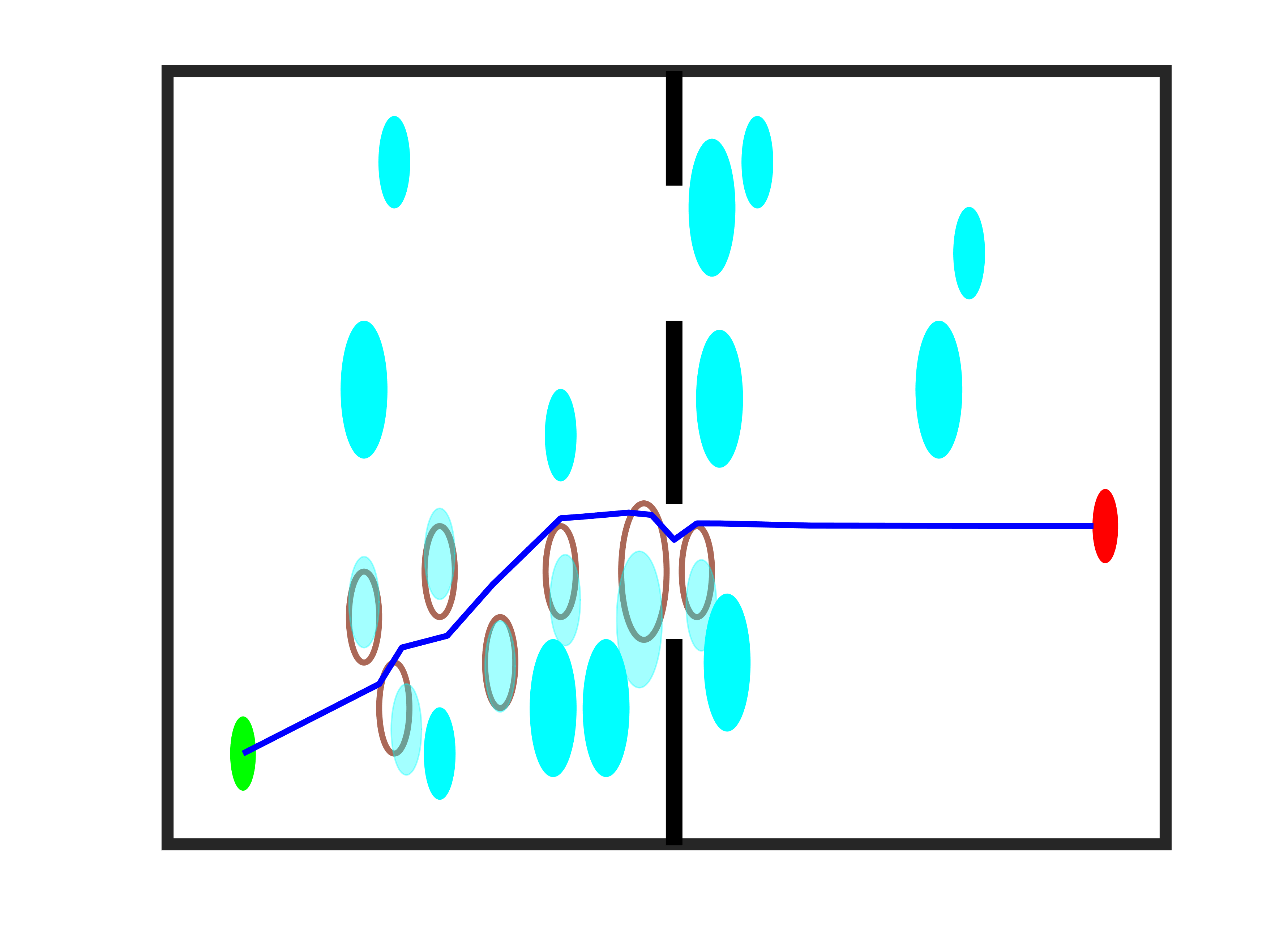}\label{fig:A2}} \\
    \subfloat[]{%
          \includegraphics[trim=50 35 30 10,clip,scale=0.3]{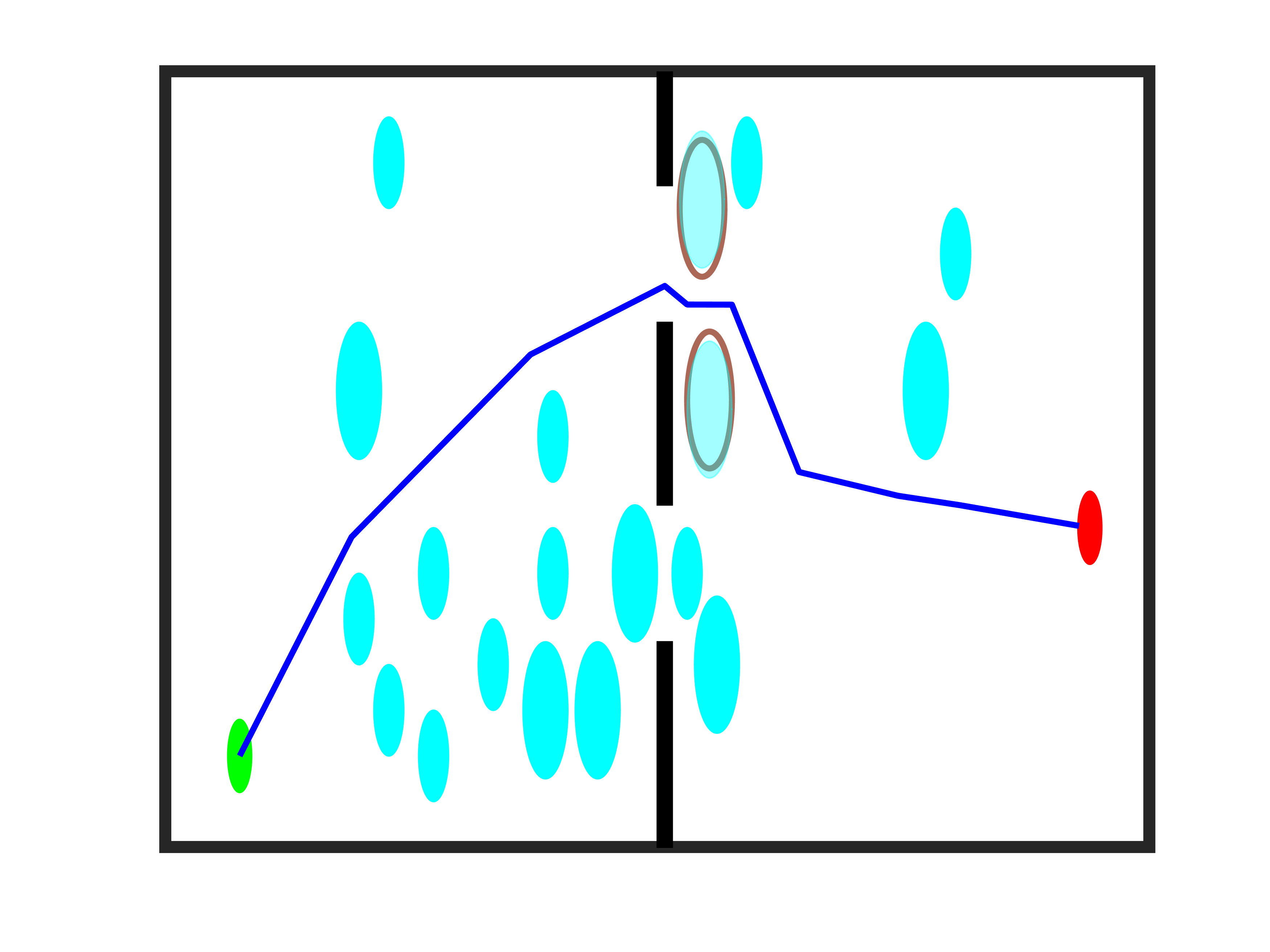}\label{fig:A3}}
    \end{tabular}}
    \end{tabular}}
    \vspace{-0.2cm}
    \caption{(a) A representative minimum obstacle displacement domain. (b)-(c) Increasing the obstacle displacement factor ($J^o$) weight from 10 to 1000, the humanoid finds a path with lower obstacle displacement but with larger path length.}
  \label{fig:application}
  \end{figure}

The evaluation domains used in this paper has been abstract. Yet, this is by no means a limitation and there exists many practical applications to the minimum obstacle displacement problem. Consider a humanoid navigating a cluttered table environment as seen in Fig.~\ref{fig:A1}. The humanoid is asked to move to the other room but its path is blocked by the tables. Employing the robot model~\eqref{eq:robot_model1} with $m=2$ and $\alpha^{re} = 1$, $\alpha^r = 1$, $\alpha^o= 10$, a 7-obstacle solution with a displacement magnitude of $4.4$ m is obtained (see Fig.~\ref{fig:A2}). Penalizing large displacements by increasing $\alpha^o$ from 10 to 1000, a longer path with a 2-obstacle solution (see Fig.~\ref{fig:A3}) is obtained with a lower displacement magnitude of $0.4$ m.
The cost $c_k$ for the last time step is the deviation of the robot location from its goal~\eqref{obj_x}. While splitting the objective function in~\eqref{eq:horizon_sliced} and~\eqref{eq:upperbound_J}, the deviation from the goal is naturally incorporated for in the last slice (sub-problem.) However, in each sub-problem the robot needs to keep track of its goal. Thus, for all the other sub-problems, we add the term $\alpha^r\alpha^{re}||\B{x}_{(i-1)T'} - \B{x}^g||^2$ with $2<=i<=m-1$. In the computation of $\epsilon$ and the values $J^r$, $J^o$ for $m>1$ in Table~\ref{table:result1}, this term is ignored to provide an exact comparison to the optimization with $m=1$. 
  
  In this work we do not deal with interacting obstacles, that is, obstacles are themselves not collision free. This can be achieved by adding the collision avoidance constraint among obstacles. However, for large obstacle number, this increases the computational burden due to addition of binary variables to achieve the linear collision avoidance constraints. Currently, we do not see how to move the obstacles and assume that the computed displacements can be achieved. For example, an obstacle might be too heavy for moving even though it gives a minimum displacement solution. This aspect may be incorporated with heavy obstacles being penalized more by giving a higher weight than the other obstacles. It is also interesting to include additional aspects such as work done by the robot which also absorbs the previous concern with force required to move the obstacles.

%% file: paper.bbl
\begin{thebibliography}{10}

\bibitem{hauser2014IJRR}
K.~Hauser, ``The minimum constraint removal problem with three robotics
  applications,'' {\em The International Journal of Robotics Research},
  vol.~33, no.~1, pp.~5--17, 2014.

\bibitem{cormen2009book}
T.~H. Cormen, C.~E. Leiserson, R.~L. Rivest, and C.~Stein, {\em Introduction to
  algorithms}.
\newblock MIT press, 2009.

\bibitem{stilman2005IJHR}
M.~Stilman and J.~J. Kuffner, ``Navigation among movable obstacles: Real-time
  reasoning in complex environments,'' {\em International Journal of Humanoid
  Robotics}, vol.~2, no.~04, pp.~479--503, 2005.

\bibitem{nieuwenhuisen2008WAFR}
D.~Nieuwenhuisen, A.~F. van~der Stappen, and M.~H. Overmars, ``An effective
  framework for path planning amidst movable obstacles,'' in {\em Algorithmic
  Foundation of Robotics VII}, pp.~87--102, Springer, 2008.

\bibitem{van2009WAFR}
J.~Van Den~Berg, M.~Stilman, J.~Kuffner, M.~Lin, and D.~Manocha, ``Path
  planning among movable obstacles: a probabilistically complete approach,'' in
  {\em Workshop on the Algorithmic Foundations of Robotics VIII, {WAFR},
  Guanajuato, Mexico}, pp.~599--614, Springer, 2009.

\bibitem{wilfong1991AMAI}
G.~Wilfong, ``Motion planning in the presence of movable obstacles,'' {\em
  Annals of Mathematics and Artificial Intelligence}, vol.~3, no.~1,
  pp.~131--150, 1991.

\bibitem{stilman2007ICRA}
M.~Stilman, J.-U. Schamburek, J.~Kuffner, and T.~Asfour, ``Manipulation
  planning among movable obstacles,'' in {\em Proceedings 2007 IEEE
  international conference on robotics and automation}, pp.~3327--3332, IEEE,
  2007.

\bibitem{dogar2011RSS}
M.~Dogar and S.~Srinivasa, ``A framework for push-grasping in clutter,'' in
  {\em Robotics: Science and systems VII}, 2011.

\bibitem{krontiris2015RSS}
A.~Krontiris and K.~E. Bekris, ``Dealing with difficult instances of object
  rearrangement.,'' in {\em Robotics: Science and Systems}, vol.~1123, 2015.

\bibitem{karami2021AIIA}
H.~Karami, A.~Thomas, and F.~Mastrogiovanni, ``{Task Allocation for Multi-robot
  Task and Motion Planning: A Case for Object Picking in Cluttered
  Workspaces},'' in {\em {AIxIA 2021 -- Advances in Artificial Intelligence}},
  (Cham), pp.~3--17, Springer International Publishing, 2022.

\bibitem{castro2013CDC}
L.~I.~R. Castro, P.~Chaudhari, J.~Tůmová, S.~Karaman, E.~Frazzoli, and
  D.~Rus, ``Incremental sampling-based algorithm for minimum-violation motion
  planning,'' in {\em 52nd IEEE Conference on Decision and Control},
  pp.~3217--3224, IEEE, 2013.

\bibitem{xu2016ASC}
B.~Xu and H.~Min, ``Solving minimum constraint removal (mcr) problem using a
  social-force-model-based ant colony algorithm,'' {\em Applied Soft
  Computing}, vol.~43, pp.~553--560, 2016.

\bibitem{krontiris2017AR}
A.~Krontiris and K.~E. Bekris, ``Trade-off in the computation of minimum
  constraint removal paths for manipulation planning,'' {\em Advanced
  robotics}, vol.~31, no.~23-24, pp.~1313--1324, 2017.

\bibitem{xu2020ICMLC}
B.~Xu, L.~Chen, and K.~Xu, ``Deep learning algorithm for minimum constraint
  removal (mcr) problem,'' in {\em Proceedings of the 2020 12th International
  Conference on Machine Learning and Computing}, pp.~52--56, 2020.

\bibitem{kaelbling2013IJRR}
L.~P. Kaelbling and T.~Lozano-P{\'e}rez, ``Integrated task and motion planning
  in belief space,'' {\em The International Journal of Robotics Research},
  vol.~32, no.~9-10, pp.~1194--1227, 2013.

\bibitem{srivastava2014ICRA}
S.~Srivastava, E.~Fang, L.~Riano, R.~Chitnis, S.~Russell, and P.~Abbeel,
  ``Combined task and motion planning through an extensible planner-independent
  interface layer,'' in {\em Robotics and Automation (ICRA), IEEE International
  Conference on}, pp.~639--646, IEEE, 2014.

\bibitem{dantam2016RSS}
N.~T. Dantam, Z.~K. Kingston, S.~Chaudhuri, and L.~E. Kavraki, ``Incremental
  task and motion planning: A constraint-based approach,'' in {\em Robotics:
  Science and Systems}, 2016.

\bibitem{garrett2018IJRR}
C.~R. Garrett, T.~Lozano-Perez, and L.~P. Kaelbling, ``{FFRob: Leveraging
  symbolic planning for efficient task and motion planning},'' {\em The
  International Journal of Robotics Research}, vol.~37, no.~1, pp.~104--136,
  2018.

\bibitem{thomas2021RAS}
A.~Thomas, F.~Mastrogiovanni, and M.~Baglietto, ``{MPTP: Motion-planning-aware
  task planning for navigation in belief space},'' {\em Robotics and Autonomous
  Systems}, vol.~141, p.~103786, 2021.

\bibitem{zhang2008WAFR}
L.~Zhang, Y.~J. Kim, and D.~Manocha, ``A simple path non-existence algorithm
  using c-obstacle query,'' in {\em Algorithmic Foundation of Robotics VII},
  pp.~269--284, Springer, 2008.

\bibitem{basch2001ICRA}
J.~Basch, L.~J. Guibas, D.~Hsu, and A.~T. Nguyen, ``Disconnection proofs for
  motion planning,'' in {\em Proceedings 2001 ICRA. IEEE International
  Conference on Robotics and Automation (Cat. No. 01CH37164)}, vol.~2,
  pp.~1765--1772, IEEE, 2001.

\bibitem{li2021RSS}
S.~Li and N.~T. Dantam, ``Learning proofs of motion planning infeasibility,''
  in {\em Robotics: Science and Systems}, 2021.

\bibitem{thomas2022IAS}
A.~Thomas and F.~Mastrogiovanni, ``{Minimum Displacement Motion Planning for
  Movable Obstacles},'' in {\em Intelligent Autonomous Systems 17}, (Cham),
  pp.~155--166, Springer Nature Switzerland, 2023.

\bibitem{hauser2013RSS}
K.~Hauser, ``Minimum constraint displacement motion planning,'' in {\em
  Proceedings of Robotics: Science and Systems IX}, (Berlin, Germany), June
  2013.

\bibitem{van2011lqg}
J.~Van Den~Berg, P.~Abbeel, and K.~Goldberg, ``Lqg-mp: Optimized path planning
  for robots with motion uncertainty and imperfect state information,'' {\em
  The International Journal of Robotics Research}, vol.~30, no.~7,
  pp.~895--913, 2011.

\bibitem{naik2018PHD}
V.~V. Naik, {\em Mixed-integer quadratic programming algorithms for embedded
  control and estimation}.
\newblock PhD thesis, IMT School for Advanced Studies Lucca, 2018.

\bibitem{Lofberg2004}
J.~L{\"{o}}fberg, ``Yalmip : A toolbox for modeling and optimization in
  matlab,'' in {\em In Proceedings of the CACSD Conference}, (Taipei, Taiwan),
  2004.

\end{thebibliography}
